\pdfoutput=1

\documentclass[11pt]{article}

\usepackage[]{acl}

\usepackage{times}
\usepackage{latexsym}

\usepackage[T1]{fontenc}

\usepackage[utf8]{inputenc}

\usepackage{microtype}

\usepackage{amsmath}
\usepackage{lipsum}
\usepackage{graphics}
\usepackage{mathtools}
\usepackage{booktabs}
\usepackage{multirow}
\usepackage{arydshln}
\usepackage{pifont}
\newcommand{\cmark}{\ding{51}}%
\newcommand{\xmark}{\ding{55}}%

\title{\texttt{R2D2}: Robust Data-to-Text with Replacement Detection}

\author{Linyong Nan \quad Lorenzo Jaime Yu Flores \quad Yilun Zhao \quad Yixin Liu \\ 
\textbf{Luke Benson} \quad \textbf{Weijin Zou} \quad \textbf{Dragomir Radev} \\
Yale University \\
  \texttt{\{linyong.nan, lj.flores, yilun.zhao\}@yale.edu}
}
\begin{document}
\maketitle
\begin{abstract}
Unfaithful text generation is a common problem for text generation systems. In the case of Data-to-Text (D2T) systems, the factuality of the generated text is particularly crucial for any real-world applications. We introduce \texttt{R2D2}, a training framework that addresses unfaithful Data-to-Text generation by training a system both as a generator and a faithfulness discriminator with additional replacement detection and unlikelihood learning tasks. To facilitate such training, we propose two methods for sampling unfaithful sentences. We argue that the poor entity retrieval capability of D2T systems is one of the primary sources of unfaithfulness, so in addition to the existing metrics, we further propose NER-based metrics to evaluate the fidelity of D2T generations. Our experimental results show that \texttt{R2D2} systems could effectively mitigate the unfaithful text generation, and they achieve new state-of-the-art results on FeTaQA, LogicNLG, and ToTTo, all with significant improvements.
\end{abstract}

\section{Introduction}
Data-to-Text generation is the task of generating a text sequence that describes some salient information of a knowledge source. Unlike Text-to-Text generation whose input source is a text sequence containing knowledge that is not extracted and represented in the canonical structured format, we assume that the input of a Data-to-Text system is represented in some structured format, e.g., RDF \cite{gardent-etal-2017-webnlg}, relational or entity tables \cite{lebret-etal-2016-neural, wiseman-etal-2017-challenges}. The Data-to-Text task can be divided into two distinct processes as in many other text generation tasks \cite{reiter_dale_2000_studies, gatt_krahmer_2018_survey}. The first process involves selecting salient information from the structured knowledge either based on natural language query or other indication of saliency, and the second process comprises organizing and planning of the previous selections to allow realization of the surface text. Although this task has been studied comprehensively in many works, from task design, modeling techniques, to application in different domains \cite{gardent-etal-2017-webnlg, lebret-etal-2016-neural, wiseman-etal-2017-challenges, novikova-etal-2017-e2e, parikh-etal-2020-totto, nan_et_al_2022_fetaqa}, existing Data-to-Text (D2T) systems exhibit a shortcoming that cannot be neglected.  
They fail to reliably generate sentences that are faithful given the salient content of the input table \cite{chen-etal-2020-logical, chen_et_al_2019_tabfact, chen-etal-2021-improving, uehara-etal-2020-learning, ji_et_al_2022_survey}. This limitation prevents the application of D2T systems in real world scenarios. We therefore need to investigate possible remedies.

We introduce a framework that prevents unfaithful Data-to-Text generation by training a Data-to-Text system both as a generator as well as a faithfulness discriminator. For faithfulness discrimination, we adopt the $\textit{replaced token detection}$ objective, which was first proposed in ELECTRA \cite{clark_et_al_2020_electra}. It was applied to the pre-training stage of the large-scale language models for more sample-efficient training of contextualized representations of sentences. ELECTRA is tasked to discriminate between original natural sentences and token-replaced sentences by locating the positions of replacement. The replaced tokens are sampled from a proposal distribution using a generator such as a Masked Language Model to fill some masked tokens. 

In our work, we perturbed the entailed reference sentences with two different methods, a knowledge-based one and a model-based one, to obtain unfaithful sentences whose surface forms are close to those of original sentences (therefore having similar sequence likelihoods), but contradict to the input table. Then we investigated ways of incorporating the discrimination task into the existing maximum likelihood learning. Specifically, we explored the settings of learning the sentence-level detection and generation in tandem, and the token-level detection and generation in tandem. In addition, we also experiment with incorporating the unlikelihood training objective \cite{welleck_et_al_2019_neural} on these unfaithful sentences to test its utility. 

We conduct experiments on three Data-to-Text datasets to test the general applicability of our approach: FeTaQA \cite{nan_et_al_2022_fetaqa}, LogicNLG \cite{chen-etal-2020-logical}, and ToTTo \cite{parikh-etal-2020-totto}.
Each dataset presents distinct challenges while faithful generation is a common problem. We find that adding the faithfulness discrimination task effectively mitigates the unfaithful Data-to-Text generation, supported by our results on multiple datasets, on all of which we are able to achieve new state-of-the-art results with evident improvements. We compare and analyze the performance of our system and existing state-of-the-art systems. To ensure the competence of comparison, we also evaluate various metrics for their aptness of faithfulness evaluation. We released our model and code at \url{https://github.com/Yale-LILY/r2d2}.

\section{Method}
\subsection{Preliminaries}
The de facto Data-to-Text task requires conditional language modeling of the sequence pair $X=(x_1,\ldots,x_M), Y=(y_1,\ldots,y_N)$ using a neural model parameterized with $\theta$: $p_\theta(y_1,\ldots,y_N|x_1,\ldots,x_M)$, where $(y_1,\ldots,y_N)$ is a natural language sentence that faithfully describes the salient part of the input data which is linearized, along with other contexts such as query or metadata, into the sequence $X=(x_1,\ldots,x_M)$. 



\begin{figure}[h]
  \centering
  \includegraphics[width=0.45\textwidth]{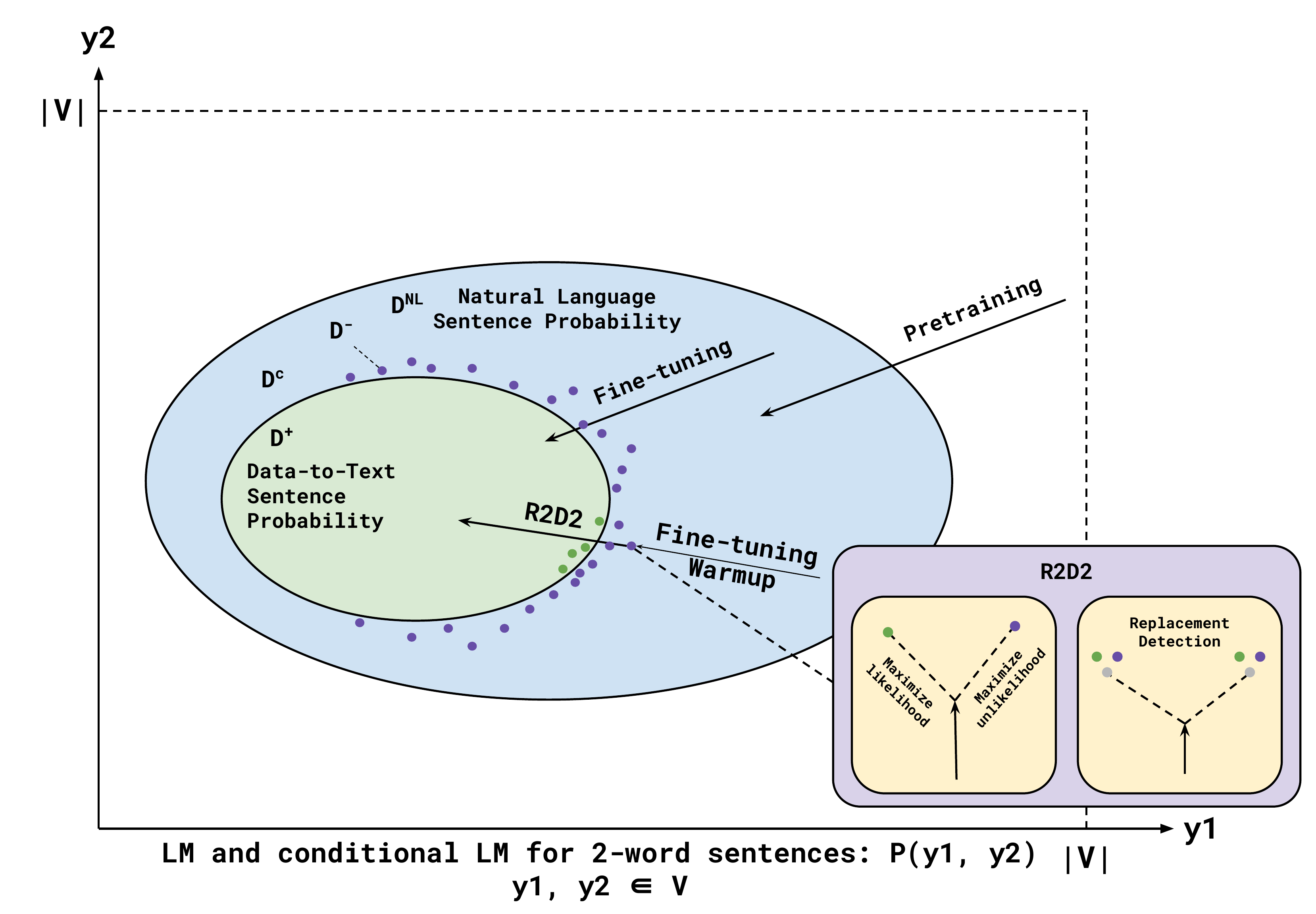}
  \caption{Illustrated R2D2 training}
  \label{fig:r2d2-finetune}
\end{figure}

We want to sample unfaithful sentences that are in the vicinity of surface forms of reference sentences, therefore also likely to be generated when only learning with maximum likelihood loss. We aim to examine the effectiveness of our proposed discrimination objectives in guiding the D2T model to attain \textbf{separable representations} for these superficially similar but factually critical sentences, and more importantly, we investigate how generation can benefit from these additional objectives for robustness. We call our method Robust Data-to-Text with Replacement Detection ($\texttt{R2D2}$), because we assign the Data-to-Text model both a generation task and a discrimination task (replacement detection). This process is illustrated in Figure \ref{fig:r2d2-finetune}.

Many existing works that study the unfaithful text generation problem in summarization and translation have investigated the source of inconsistencies between the system output and the input \cite{cao_et_al_2017_faithful, maynez-etal-2020-faithfulness, goyal-durrett-2020-evaluating, goyal-durrett-2021-annotating, chen-etal-2021-improving}. The main source of unfaithfulness is that the outputs contain facts that cannot be entailed (necessary consequence) from any explicitly stated facts or inferences derived from the input table. Since facts can be represented using subjects, predicates, and objects, these contradictions originate from wrong predictions of entities (subject or object), predicates, or wrong arrangements. This motivates our proposal of different methods of obtaining unfaithful sentences in Section \ref{replacement}. Then we describe two learning objectives that we proposed to add to the Data-to-Text modeling: 1) replacement detection objective in Section \ref{replacement-detection}; and 2) unlikelihood objective in Section \ref{unlikelihood}. In Section \ref{r2d2-finetuning}, we formulate our \texttt{R2D2} fine-tuning that leverages these two objectives in addition to the standard negative log likelihood loss for robust training of a Data-to-Text model. We also propose NER-based evaluation metrics to complement the existing evaluations of faithfulness in Section \ref{ner-evaluation}.

\begin{figure*}[h]
  \centering
  \includegraphics[width=0.95\textwidth]{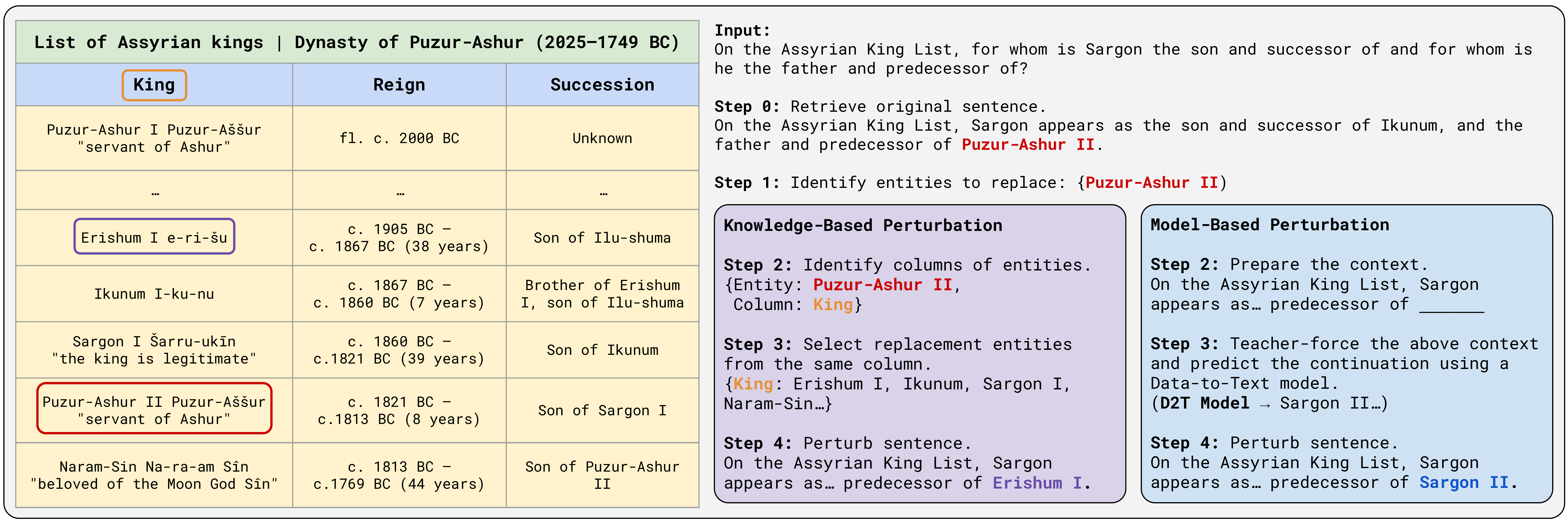}
  \caption{Faithfulness-based replacement method}
  \label{fig:replacement-method}
\end{figure*}

\subsection{Faithfulness-based Replacement}
\label{replacement}
We aim to obtain sentences that are not entailed from the input table by replacing tokens in the original sentences. As we discussed in the previous paragraph, one of the major sources of unfaithfulness is the erroneous prediction of entities or predicates. It is challenging to reliably extract the predicates in canonical form, compare those in the system's generations to those in the input table, and find replacements for them. However, it is feasible to extract and compare entities from the system generations and the input table. Therefore we replace a span of tokens that constitute an entity in the original sentence with another candidate entity comprising one or more tokens. We adopted a RoBERTa-large-based named entity recognizer\footnote{\url{https://huggingface.co/Jean-Baptiste/roberta-large-ner-english}} to extract all the entities in a sentence. To sample one unfaithful sentence, we select one of these entities, with those located near the end of the sentence to have a higher probability of being selected. This way, for the token-level detection task, the model has more chances of obtaining a reasonable amount of context for performing discrimination. We propose two different methods of determining the candidate replacement, which are illustrated in Figure \ref{fig:replacement-method} and described next. 

\paragraph{Knowledge-based Method}
Entities of similar types are usually the suspects of the wrong predictions. Identifying such entities from the input table is nontrivial in other text generation tasks whose input table is unstructured text. In D2T, the source arranges similar entities together. For example, when the input is a table, similar entities can be accessed in the same column; when the input is semantic triple set, we can obtain similar entities by searching those applicable to certain predicates. Our knowledge-based replacement method exploits the structure of the input table to retrieve similar entities, replacing the original entity with one in the sentence will lead to contradiction. This process is illustrated in Figure \ref{fig:replacement-method}. Note that for the datasets we experimented on, we assume the entity to be replaced is the only choice to make the sentence entailed, and replacement of any other entities shown in the input table will violate the faithfulness of sentences. This assumption does not necessarily hold in some examples, on which we will elaborate in Section \ref{analysis}.

\paragraph{Model-based Method}
Another way to obtain replacement candidates is by sampling from a proposal distribution, in a way similar to ELECTRA. We sample replacements from the baseline Data-to-Text model by teacher forcing partial sentence up to the entity that needs to be replaced. Next we collect the D2T model predictions of the continuation with nucleus sampling \cite{holtzman_2020_degeneration}, followed by extracting altered entities shown in the predicted continuations to determine the replacement candidates. With this approach, we are able to expose and further train the D2T model with errors of its own predictions.

\begin{figure*}[htp]
\begin{equation}
\label{eq:1}
    \resizebox{0.6\hsize}{!}{
        $\displaystyle\mathcal{L}_{\text{RD}_{\text{sent}}}(X^{(i)}, Y) = - \left[ l \cdot \log p(l|Y, X^{(i)}) + (1-l) \cdot \log \left(1-p(l|Y, X^{(i)})\right)\right]$
    }
\end{equation}
\begin{equation}
\label{eq:2}
    \resizebox{0.7\hsize}{!}{
        $\displaystyle\mathcal{L}_{\text{RD}_{\text{token}}}(X^{(i)}, Y) = - \left[ \sum_{t=1}^{|Y|} l_t \cdot \log p(l_t|y_{\leq t}, X^{(i)}) + (1-l_t) \cdot \log \left(1-p(l_t|y_{\leq t}, X^{(i)})\right)\right]$
    }
\end{equation}
\begin{equation}
\label{eq:3}
    \resizebox{0.9\hsize}{!}{
        $\displaystyle\mathcal{L}_{\text{UL}}(X^{(i)}, Y_{\text{False}}^{(i, j)}) = -\left[\sum_{t=\{t|y_t\in \mathcal{C}^{(i,j)}\}}^{|\mathcal{C}^{(i, j)}|} y_t\cdot \log (1 - \hat{y_t}) + (1-y_t)\cdot \log(\hat{y_t})+\sum_{t=\{t|y_t\notin \mathcal{C}^{(i, j)}\}}^{|Y_{\text{False}}^{(i, j)}\setminus \mathcal{C}^{(i, j)}|} y_t\cdot \log (\hat{y_t}) + (1-y_t)\cdot \log (1-\hat{y_t})\right]$
    }
\end{equation}
\begin{equation}
\label{eq:4}
    \resizebox{0.5\hsize}{!}{
        $\displaystyle\mathcal{L}_{\text{NLL}}(X^{(i)}, Y_{\text{True}}^{(i)}) =  - \sum_{t=1}^{|Y_\text{True}^{(i)}|} y_t\cdot \log (\hat{y_t}) +     (1-y_t)\cdot \log (1-\hat{y_t})$
    }
\end{equation}
\begin{equation}
\label{eq:5}
    \resizebox{0.94\hsize}{!}{
        $\displaystyle\mathcal{L}_{\text{R2D2}} = \sum_{i=1}^{|\mathcal{D}|}\frac{1}{|N^{(i)}|+1}\left[\lambda \left(\mathcal{L}_\text{NLL}(X^{(i)}, Y_{\text{True}}^{(i)}) + \sum_{j=1}^{|N^{(i)}|} \mathcal{L}_\text{UL}(X^{(i)}, Y_{\text{False}}^{(i, j)}) \right)+ (1-\lambda)\left(\mathcal{L}_{\text{RD}_{\{\text{sent}, \text{token}\}}}(X^{(i)}, Y_{\text{True}}^{(i)})+\sum_{j=1}^{|N^{(i)}|}\mathcal{L}_{\text{RD}_{\{\text{sent}, \text{token}\}}}(X^{(i)}, Y_{\text{False}}^{(i,j)})\right)\right]$
    }
\end{equation}
\end{figure*}

\subsection{Replacement Detection in Generation}
\label{replacement-detection}
For each entailed sentence $Y_{\text{True}}^{(i)}$ that we sample from the original dataset $\mathcal{D}$ where $i=1,\ldots,|\mathcal{D}|$, we generate $N^{(i)}$ contradictory sentences which we denote as $Y_{\text{False}}^{(i,j)}$ for $j=1,\ldots, |N^{(i)}|$. The number of contradictory sentences we can generate given an entailed sentence depends on the number of entities found in the original sentence, the input, and the replacement method applied. 

\begin{figure}[h]
  \centering
  \includegraphics[width=0.45\textwidth]{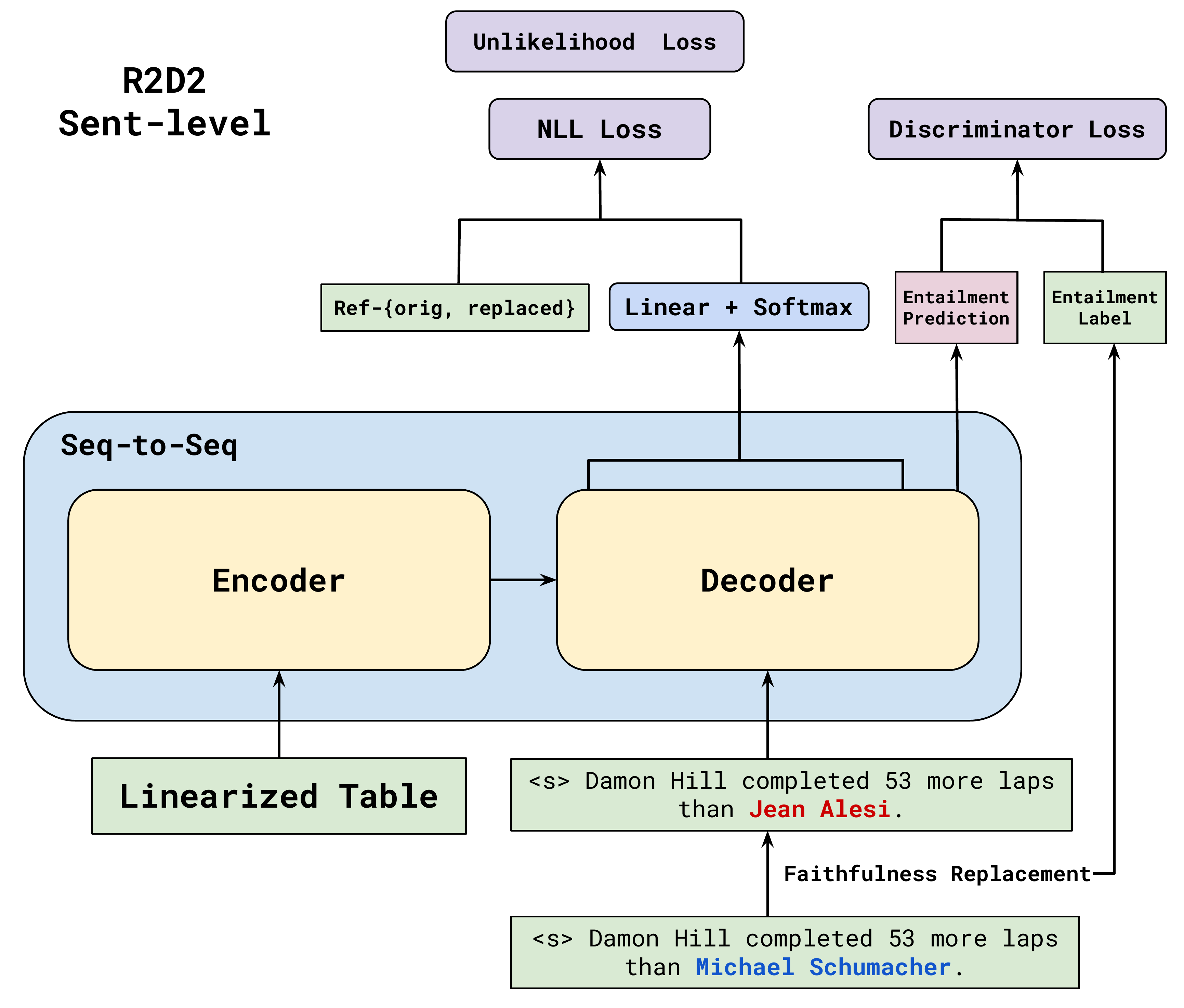}
  \caption{R2D2 sentence-level architecture}
  \label{fig:r2d2-sent-level}
\end{figure}

As shown in Figure \ref{fig:r2d2-sent-level}, we add a sentence-level replacement detection task to the existing Sequence-to-Sequence framework by eliciting the decoder to generate a probability of the teacher-forced sentence being entailed or contradictory  at the end of the generation, similar to the sequence classification usage of BART \cite{lewis-etal-2020-bart}, except that in BART, the same sequence that needs to be classified is fed into both the encoder and decoder. The loss for sentence-level replacement detection is defined by Equation \eqref{eq:1}.

\begin{figure}[h]
  \centering
  \includegraphics[width=0.45\textwidth]{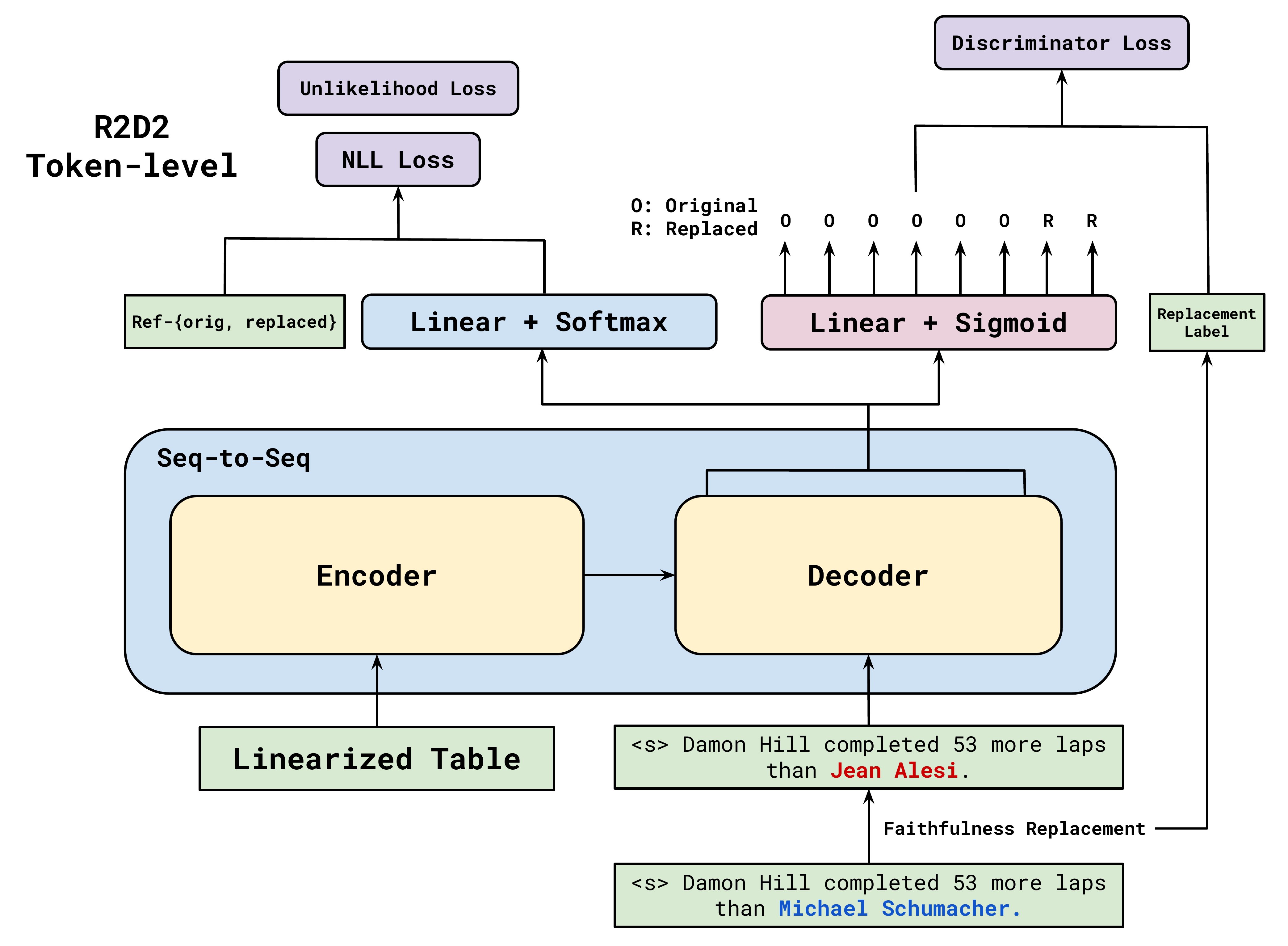}
  \caption{R2D2 token-level architecture}
  \label{fig:r2d2-token-level}
\end{figure}

A more challenging task is to perform a fine-grained, token-level discrimination, as shown in Figure \ref{fig:r2d2-token-level}. Instead of predicting a discrimination probability at the end of generation, we task the decoder to perform discrimination at every step of token generation. Specifically, we use the per-step last hidden output of the decoder, which encodes the source contexts and teacher-forced partial generation contexts, to compute the discrimination probability with a linear and sigmoid layer. The token-level replacement detection loss is defined by Equation \eqref{eq:2}.

\subsection{Replacement Unlikelihood Training}

\label{unlikelihood}
Unlikelihood training is first proposed in \cite{welleck_et_al_2019_neural} to address the repetition problem of the neural text generation. This training objective aims to decrease the decoder's probability of generating tokens that are already seen in the teacher-forced generation contexts. The applicability of this objective to the Data-to-Text task is also argued in \cite{uehara-etal-2020-learning}. 

Instead of using the generated tokens to construct the negative candidate set defined for each step, we define the sentence-level negative candidate span $\mathcal{C}^{(i,j)}$ for each contradictory sentence $Y_{\text{False}}^{(i,j)}$. The span contains, for each time step, one replaced token that should have a low probability of being generated. We calculate the sequence-level unlikelihood loss for this replaced token span and apply regular likelihood loss for other original tokens, which we denote as $Y_{\text{False}}^{(i, j)}\setminus \mathcal{C}^{(i, j)}$. We denote the per step prediction as $\hat{y_t} = p(y_t|y_{<t}, X^{(i)})$. The unlikelihood loss for the entire sentence is specified in Equation \eqref{eq:3}.

\subsection{\texttt{R2D2} Fine-tuning}
\label{r2d2-finetuning}

We propose the final \texttt{R2D2} fine-tuning loss objective as in equation \eqref{eq:5}. It combines a generation task loss component and a discrimination task loss component. Each of them is calculated from one entailed instance and $N^{(i)}$ contradictory instances. The generation task component consists of a regular negative log likelihood loss for the entailed instance, as described in Equation \eqref{eq:4}, and an unlikelihood loss for the contradictory instances. The discrimination task component contains either sentence-level or token-level replacement detection loss for both entailed and contradictory instances. We use $\lambda$ to balances the importance between the two loss components.

\subsection{NER-based Evaluation Metrics}
\label{ner-evaluation}
To better understand how the D2T system's retrieval capability correlates with faithfulness, we propose information-extraction based metrics that compare the named entities contained in the generated sentences to those contained in the reference sentences or input data. We believe these metrics help us better distinguish between the unfaithfulness caused by wrong retrieval of entities and that caused by wrong prediction of predicates/relations. Specifically, we propose the following five indicators:
\begin{itemize}
  \item \textbf{Reference coverage (RC)}: percentage of entities in the reference that are also shown in the prediction.
  \item \textbf{Ref-hit \& Input-hit (RI)}: percentage of entities shown in the prediction that are shown in both the reference and input table.
  \item \textbf{Ref-hit \& Input-miss (RM)}: percentage of predicted entities that are shown in the reference but not the input table. This case is rare since it indicates the existence of entities that are not input-grounded in the reference.
  \item \textbf{Ref-miss \& Input-hit (MI)}: percentage of predicted entities that are shown in the table, but not in the reference. This case identifies wrong or unnecessary retrieval of entities from the input (when the indication of saliency is evident). 
  \item \textbf{Ref-miss \& Input-miss (MM)}: percentage of predicated entities that are neither shown in the table nor the reference. This case identifies the prediction of entities likely by hallucination.
\end{itemize}

\begin{figure}[h]
  \centering
  \includegraphics[width=0.45\textwidth]{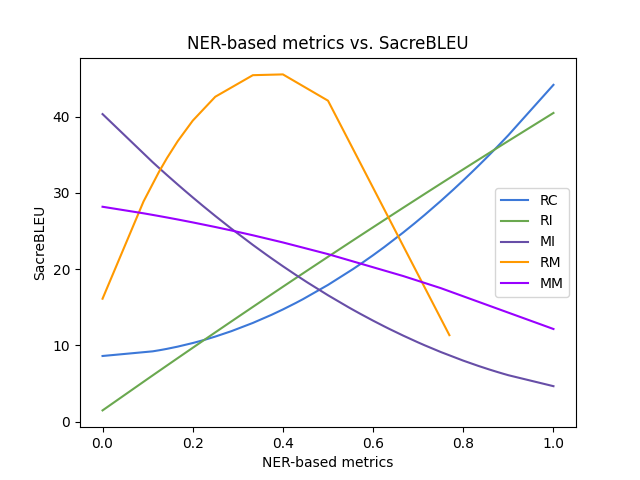}
  \caption{NER-based metrics vs sacreBLEU}
  \label{fig:ner-metrics-sacrebleu-correlation}
\end{figure}

\begin{figure}[h]
  \centering
  \includegraphics[width=0.45\textwidth]{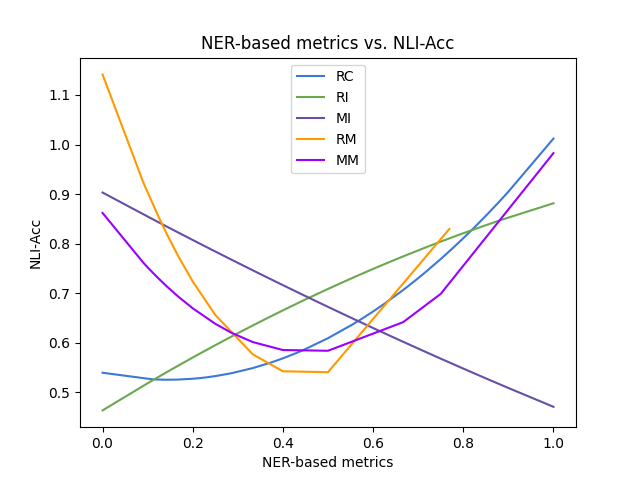}
  \caption{NER-based metrics vs NLI-Acc}
  \label{fig:ner-metrics-nliacc-correlation}
\end{figure}

Figure \ref{fig:ner-metrics-sacrebleu-correlation} and Figure \ref{fig:ner-metrics-nliacc-correlation} show the correlation between NER-based metrics and sacreBLEU, NLI-Acc, respectively. As expected, the reference coverage rate positively correlates with both metrics. While both reference hit and input hit are important, the rate of predicted entities not shown in reference negatively correlates with sacreBLEU (MI and MM) and NLI-Acc (MI). The trend is less clear for RM and MM since these are rare cases, which can also be shown in Table \ref{tab:ner-evaluation}.

Since many existing automatic metrics for text generation tasks are not proposed with an aim of reflecting the faithfulness of the sentences, an examination of all metrics reported in our work is crucial for interpreting the results. As we are able to reliably generate an unfaithful version of most of the reference texts, we contaminate the references in the FeTaQA test split in a controlled manner: we generate five variants of texts with different percentage of the references being replaced with their unfaithful parallel ($0\%$ version contains only the references and $100\%$ version contains only the unfaithful sentences). We evaluate the variants that are contaminated to different degrees using the evaluation metrics we reported, in order to investigate how reliable they are in reflecting the faithfulness of any system generated sentences.

\begin{table*}[h]
\centering
\resizebox{\hsize}{!}{
\begin{tabular}{@{\extracolsep{3pt}}ccccccccccccc@{}}
\toprule
\multirow{2.5}{*}{\textbf{\begin{tabular}[c]{@{}c@{}}\% Unfaithful\\ Sentences\end{tabular}}} & \textbf{Fact-based} & \multicolumn{6}{c}{\textbf{String-based}} & \multicolumn{5}{c}{\textbf{NER-based}} \\
\cmidrule{2-2} \cmidrule{3-8} \cmidrule{9-13} 
 & \textbf{NLI-Acc} & \textbf{sacreBLEU} & \textbf{Rouge-1/2/L} & \textbf{PARENT} & \textbf{TER} & \textbf{METEOR} & \textbf{BERTScore} & \textbf{RC} & \textbf{RI} & \textbf{RM} & \textbf{MI} & \multicolumn{1}{l}{\textbf{MM}} \\
\midrule
\textbf{0\% (ref.)} & 86.17 & 100.0 & 100/100/100 & 89.87 & 0.0 & 99.99 & 100.0 & 99.95 & 94.15 & 5.80 & 0.0 & 0.05 \\
\textbf{25\%} & 73.29 & 96.1 & 97.62/96.31/97.62 & 87.23 & 2.57 & 98.08 & 99.43 & 96.37 & 90.82 & 5.67 & 2.98 & 0.53 \\
\textbf{50\%} & 61.21 & 92.2 & 95.28/92.63/95.29 & 84.63 & 5.18 & 96.24 & 98.87 & 93.01 & 87.53 & 5.55 & 5.86 & 1.06 \\
\textbf{75\%} & 48.63 & 88.3 & 92.87/88.87/92.87 & 81.91 & 7.84 & 94.36 & 98.29 & 89.32 & 84.33 & 5.31 & 8.81 & 1.54 \\
\textbf{100\%} & 36.20 & 84.4 & 90.47/85.13/90.48 & 79.16 & 10.36 & 92.48 & 97.72 & 85.70 & 80.93 & 5.24 & 11.72 & 2.11 \\
\bottomrule
\end{tabular}
}%
\caption{Reliability test result of evaluation metrics for detecting unfaithfulness}
\label{tab:metric-test}
\end{table*}

As shown in Table \ref{tab:metric-test}, most of the metrics are able to reflect the degree of unfaithfulness contained in the prediction texts, though our test only contains the type of unfaithfulness that originates from erroneous selection of entities. A more rigorous study would test other types of unfaithfulness, such as wrong prediction of relations or arrangement of entities. Nevertheless, we observe that some metrics, especially NLI-Acc, are more sensitive to the type of unfaithfulness that we tested, while an unfaithful sentence can still obtain a very high BERTScore \cite{zhang-etal-2020-bertscore}.

\section{Experiments}
We first introduce the datasets that we experiment with in Section \ref{datasets}, and the metrics we adopted for evaluation in Section \ref{metrics}. Then we report the baseline models we are comparing with, and the implementation and training details in Section \ref{exp-setting}. In Section \ref{auto-eval} and \ref{human-eval}, we report both the automatic evaluation results and the human evaluation results, followed by an analysis of them in Section \ref{analysis}. 

\subsection{Datasets}
\label{datasets}

\noindent \texttt{FeTaQA} \cite{nan_et_al_2022_fetaqa} is a free-form table question answering dataset. It introduces a task that requires retrieving the correct contents from the table based on the question, integrating and inferring from the retrieved facts, and generating a free-form answer. A sentence that contains erroneous selections of facts, even if they appear in the input table, are still considered as unfaithful, because the sentence is inconsistent (incorrect) with the input question. 

\noindent \texttt{LogicNLG} \cite{chen-etal-2020-logical} is a table-to-text dataset that tasks the the models to generate logically entailed sentences. There is no indication of what is considered as salient given a table, as long as the generated sentence is entailed given all the facts contained in the table. Since there are numerous entailed facts that are different from the references in the surface-form, they propose input-based metrics that compare the facts in the generated sentence and those in the input. Although we report the experiment results of reference-based metrics for this dataset, it is worth noting that unlike other datasets mentioned above, faithfulness to the input is more important for LogicNLG then faithfulness to the references.

\noindent \texttt{ToTTo} \cite{parikh-etal-2020-totto} is a table-to-text dataset that contains annotations of salient content of tables (highlighted table cells). The task does not require any content selection (when only highlighted cells constitute the input), but only text planning and surface realization of the inputs, which are expected to be described with full coverage, and the median number of highlighted cells is 3.


\subsection{Evaluation Metrics}
\label{metrics}
We report results of a variety of automatic evaluation metrics used in the past studies to provide a comprehensive comparison of existing methods and our proposed method. We include fact-verification based metrics, NLI-Acc \cite{chen-etal-2020-logical, chen_et_al_2019_tabfact}, which specifically aims to evaluate the faithfulness of the sentences. We also report string-based metrics that evaluate the $n$-gram or other types of string match between predictions and references, such as sacreBLEU \cite{post-2018-call}, ROUGE-\{1, 2, L\} \cite{lin-2004-rouge}, TER \cite{snover-etal-2006-study}, METEOR \cite{banerjee-lavie-2005-meteor}, PARENT \cite{dhingra-etal-2019-handling} (which also leverages the input data). In addition, we report evaluation results of the NER-based metrics.

\subsection{Experiment Settings}
\label{exp-setting}
\paragraph{Baselines}
The state-of-the-art system for the Data-to-Text task is obtained by fine-tuning a pre-trained sequence-to-sequence model, such as T5 \cite{raffel-etal-2020-t5}, BART \cite{lewis-etal-2020-bart} or Longformer-Encoder-Decoder (LED) \cite{beltagy-etal-2020-longformer}. We report evaluations of existing state-of-the-art models for all datasets. We also fine-tuned T5 on FeTaQA, LogicNLG and ToTTo 
by ourselves, so that the learning objective is the key control variable in our comparison.

\paragraph{Implementation}
We use T5-base 
as the pre-trained checkpoint from which we fine-tune either regularly (Reg-FT) or using our proposed method (\texttt{R2D2}-FT). For \texttt{R2D2}-FT, we initialize our model from a checkpoint that has been fine-tuned regularly for 15 epochs, and train the additional linear layer for sentence or token replacement detection from random initialization. We find this fine-tuning warmup help improve the performance in general. We use the Adafactor optimizer \cite{shazeer-stern-2018-adafactor} 
with a learning rate of \texttt{5e-5}. We use the batch size of 8 for FeTaQA and 32 for the others.

\paragraph{\texttt{R2D2} Configuration}
To assess the necessity of the discrimination loss and unlikelihood loss, we experimented fine-tuning T5 only with the discrimination loss, or only with the unlikelihood loss, or both (all in addition to the NLL loss). For discrimination loss, we also experiment with adding sentence-level or token-level discrimination to investigate the effect of discrimination granularity in assisting faithful text generation. For all the training variants above, we also compare two methods of obtaining the contradictory sentences, knowledge-based and model-based methods. Since the number of contradictory sentences obtained (which we denote as $N^{(i)}$ in Section \ref{r2d2-finetuning}) varies depending on the method used (as shown in Table \ref{tab:perturb-size}), we also experiment with using different numbers of contradictory sentences in the \texttt{R2D2} fine-tuning: $N^{(i)} = 1$ (\texttt{xsmall}), $3$ (\texttt{small}), $5$ (\texttt{medium}), $10$ (\texttt{large}) or max (\texttt{full}). Since the maximum size of the perturbations obtained by model-based method is \texttt{small}, we only compared \texttt{xsmall} and \texttt{full} for the model-based method setting. 


\begin{table}[]
\centering
\resizebox{\hsize}{!}{
\begin{tabular}{@{}cccc@{}}
\toprule
\textbf{Dataset}  & \begin{tabular}[c]{@{}c@{}}\textbf{Original}\\ \textbf{Train-split}\\ \textbf{Size}\end{tabular} & \textbf{Knowledge-based}  & \textbf{Model-based}    \\ 
\midrule
\textbf{FeTaQA}   & 7,325  & 238,891 ($\times$32.6)  & 25,313 ($\times$3.5)  \\
\textbf{LogicNLG} & 21,873  & 413,247 ($\times$18.9)   & 91,183 ($\times$4.2)  \\
\textbf{ToTTo}    & 120,761  & 1,165,067 ($\times$9.6) & 450,770 ($\times$3.7) \\
\bottomrule
\end{tabular}
}%
\caption{Number of obtainable contradictory sentences for the train-split of all datasets.}
\label{tab:perturb-size}
\end{table}


\begin{table*}[h]
\centering
\resizebox{0.9\textwidth}{!}{
\begin{tabular}{@{\extracolsep{3pt}}ccccccc@{}}
\toprule
\multirow{2.5}{*}{\textbf{Systems}} & \textbf{Fact-based} & \multicolumn{5}{c}{\textbf{String-based}} \\
\cmidrule{2-2} \cmidrule{3-7} 
 & \multirow{1}{*}{\textbf{NLI-Acc}} & \multirow{1}{*}{\textbf{sacreBLEU}} & \multirow{1}{*}{\textbf{Rouge-1/2/L}} & \multirow{1}{*}{\textbf{PARENT}} & \multirow{1}{*}{\textbf{TER}} & \multirow{1}{*}{\textbf{METEOR}}  \\
 \midrule \noalign{\vskip 0.3ex}
\citet{xie_et_al_2022_unifiedskg} & - & 29.9 & 61.77/39.44/51.93 & - & - & 48.53 \\
\noalign{\vskip 0.7ex}\hdashline\noalign{\vskip 0.7ex}
\textbf{Reg-FT} & 74.79 & 30.5 & 63.47/41.77/54.04 & 44.24 & 66.37 & 55.2  \\
\textbf{\texttt{R2D2}-FT} & \textbf{77.93} & \textbf{31.5} & \textbf{63.50/41.71/54.05} & \textbf{45.32} & 68.55 & \textbf{56.27}  \\
\bottomrule
\end{tabular}
}%
\caption{FeTaQA-test automatic evaluation result}
\label{tab:fetaqa-auto-sc}
\end{table*}


\begin{table*}[h]
\centering
\resizebox{\textwidth}{!}{
\begin{tabular}{@{\extracolsep{3pt}}ccccccccc@{}}
\toprule
\multirow{2.5}{*}{\textbf{Systems}} & \multicolumn{2}{c}{\textbf{Fact-based}} & \multicolumn{6}{c}{\textbf{String-based}} \\
\cmidrule{2-3} \cmidrule{4-9}
 & \multirow{1}{*}{\textbf{NLI-Acc}} & \multirow{1}{*}{\textbf{SP-Acc}} & \multirow{1}{*}{\textbf{BLEU-1/2/3}} & \multirow{1}{*}{\textbf{sacreBLEU}} & \multirow{1}{*}{\textbf{Rouge-1/2/L}} & \multirow{1}{*}{\textbf{PARENT}} & \multirow{1}{*}{\textbf{TER}} & \multirow{1}{*}{\textbf{METEOR}}  \\
 \midrule
\citet{chen-etal-2021-de} & 76.9 & 43.9 & 49.5/28.6/15.3 & - & - & - & - & - \\
\noalign{\vskip 0.7ex}\hdashline\noalign{\vskip 0.7ex}
\textbf{Reg-FT} & 84.11 & 45.97 & 51.63/32.24/18.75 & 18.2 & 42.74/20.89/36.77 & 32.36 & 86.38 & 36.55 \\
\textbf{\texttt{R2D2}-FT} & \textbf{85.57} & \textbf{50.80} & \textbf{51.76/32.42/18.65} & \textbf{18.5} & 42.63/20.73/36.84 & 31.38 & \textbf{80.97} & 35.73 \\
\bottomrule
\end{tabular}
}%
\caption{LogicNLG-test automatic evaluation result}
\label{tab:logicnlg-auto}
\end{table*}


\begin{table*}[h]
\centering
\resizebox{\hsize}{!}{
\begin{tabular}{@{\extracolsep{5pt}}cccccccc}
\toprule
\multirow{2.5}{*}{\textbf{Systems}} & \multirow{2.5}{*}{\textbf{Split}} & \textbf{Fact-based} & \multicolumn{5}{c}{\textbf{String-based}} \\
\cmidrule{3-3} \cmidrule{4-8}
 & & \multirow{1}{*}{\textbf{NLI-Acc}} & \multirow{1}{*}{\textbf{sacreBLEU}} & \multirow{1}{*}{\textbf{Rouge-1/2/L}} & \multirow{1}{*}{\textbf{PARENT}} & \multirow{1}{*}{\textbf{TER}} & \multirow{1}{*}{\textbf{METEOR}} \\
\midrule
 & All & - & 47.7 & - & 57.1 & - & -  \\
 & Overlap & - & - & - & - & - & - \\
\multirow{-3}{*}{\citet{kale-rastogi-2020-text}} & Nonoverlap & - & 39.6 & - & 52.6 & - & -  \\
\noalign{\vskip 0.7ex}\hdashline\noalign{\vskip 0.7ex}
 & All & 90.29 & 48.7 & 75.92/55.99/67.40 & 58.43 & 48.53 & 71.01 \\
 & Overlap & 91.17 & 56.6 & 79.95/62.39/72.78 & 62.88 & 40.85 & 75.56 \\
\multirow{-3}{*}{\textbf{Reg-FT}} & Nonoverlap & 89.42 & 41.0 & 72.04/49.80/62.20 & 54.14 & 55.69 & 66.54  \\
\noalign{\vskip 0.7ex}\hdashline\noalign{\vskip 0.7ex}
 & All & \textbf{91.27} & \textbf{49.2} & 75.54/55.59/67.06 & \textbf{59.05} & 50.40 & \textbf{71.99}  \\
 & Overlap & \textbf{91.70} & \textbf{56.7} & 79.42/61.68/72.14 & \textbf{62.89} & 42.72 & \textbf{76.47}  \\
\multirow{-3}{*}{\textbf{\texttt{R2D2}-FT}} & Nonoverlap & \textbf{90.86} & \textbf{41.9} & 71.79/49.69/62.15 & \textbf{55.35} & 57.57 & \textbf{67.67} \\
\bottomrule
\end{tabular}
}%
\caption{ToTTo-dev automatic evaluation result}
\label{tab:totto-auto}
\end{table*}


\begin{table}[]
\centering
\resizebox{\hsize}{!}{
\begin{tabular}{@{}cccccccc@{}}
\toprule
\textbf{Dataset} & \textbf{Systems} & \textbf{Split} & \textbf{RC} & \textbf{RI} & \textbf{RM} & \textbf{MI} & \textbf{MM} \\
\midrule
\multirow{2}{*}{\textbf{FeTaQA}} & \textbf{Reg} & A & 72.30 & 69.88 & 1.48 & 24.90 & 3.73 \\
 & \textbf{\texttt{R2D2}} & A & \textbf{73.06} & \textbf{70.92} & \textbf{1.62} & \textbf{23.82} & \textbf{3.64} \\
 \noalign{\vskip 0.7ex}\hdashline\noalign{\vskip 0.7ex}
 \multirow{2}{*}{\textbf{LogicNLG}} & \textbf{Reg} & A & 37.29 & 26.07 & 0.97 & 61.60 & 11.36 \\
 & \textbf{\texttt{R2D2}} & A & \textbf{37.93} & \textbf{26.62} & \textbf{0.44} & \textbf{67.12} & \textbf{5.82} \\
 \noalign{\vskip 0.7ex}\hdashline\noalign{\vskip 0.7ex}
\multirow{6}{*}{\textbf{ToTTo}} & \multirow{3}{*}{\textbf{Reg}} & A & 82.47 & 74.95 & 3.14 & 17.66 & 4.24 \\
 &  & O & 84.87 & 77.95 & 3.40 & 15.10 & 3.54 \\
 &  & N & 80.14 & 72.04 & 2.89 & 20.14 & 4.92 \\
 & \multirow{3}{*}{\textbf{\texttt{R2D2}}} & A & \textbf{83.24} & 74.06 & 3.26 & 18.31 & 4.37 \\
 &  & O & \textbf{85.45} & 76.90 & 3.63 & 15.70 & 3.77 \\
 &  & N & \textbf{81.08} & 71.30 & 2.89 & 20.85 & 4.95 \\
 \bottomrule
\end{tabular}
}%
\caption{NER-based automatic evaluation result. A, O, N stands for All, Overlap and Nonoverlap.}
\label{tab:ner-evaluation}
\end{table}

\subsection{Automatic Evaluation}
\label{auto-eval}
We report the performance of the previous state-of-the-art system, T5 fine-tuned only with negative log likelihood (NLL) loss by ourselves, and the best T5 fine-tuned with \texttt{R2D2} loss for FeTaQA (Table \ref{tab:fetaqa-auto-sc}), LogicNLG (Table \ref{tab:logicnlg-auto}) and ToTTo (Table \ref{tab:totto-auto}), based on metrics used in the existing literature. In Table \ref{tab:ner-evaluation}, we also report their performances using the NER-based metrics that we proposed. We also report the full experiment results of FeTaQA that contain evaluations of different \texttt{R2D2} configurations in Table \ref{tab:full-result} of the Appendix. 

We obverse that across all the datasets, most of the systems fine-tuned with different \texttt{R2D2} configurations are able to perform better than system that is fine-tuned only with NLL loss. As expected, the improvements are more evident in the fact-verification-based metrics that evaluate the faithfulness of the sentences. We find that the best \texttt{R2D2} configuration requires both the token-level discrimination and the unlikelihood learning objectives, and that contradictory sentences obtained by knowledge-based perturbation are more beneficial than those obtained by model-based perturbation for fine-tuning when all other configuration constants are controlled. We also find that system performance does not necessarily improve as we increase the number of unfaithful sentences used for fine-tuning, and that the best configuration for $N^{(i)}$ seems to be three to five in most cases. 

We examine the improvement of faithfulness using the NER-based metrics, and find that the coverage of the entities appeared in the reference (RC) improves across all datasets. For FeTaQA, we notice that our system is able to retrieve \textit{input-grounded entities} more accurately (shown by increased RI and decreased MI scores). For LogicNLG which has no right or wrong retrieval of \text{input-grounded entities} as long as the description of them is faithful, we obverse an evident decline of MM and increments of both RI and MI (with more evident gain in MI), indicating that our system is able to reduce hallucinations of irrelevant entities and instead retrieving input-grounded ones.

\subsection{Human Evaluation}
\label{human-eval}
Since the automatic evaluations are not always reliable in determining the faithful aspect of a sentence, which can be seen in our metrics reliability test shown in Table \ref{tab:metric-test}: around 14\% faithful sentences are deemed to be unfaithful by the NLI-Acc metric, and more importantly, it fails to identify around 36\% of the unfaithful sentences. We conduct the human evaluation based on two criteria: a sentence is (1) \textit{faithful} if \textbf{all facts} contained are entailed by the input, and when a question is present in the input, the sentence only contains \textbf{necessary facts}; (2) \textit{adequate with respect to reference} if the sentence contains \textbf{same or more facts} than the reference. We asked three human evaluators to evaluate 200 samples of each dataset (100 samples in each of the overlap/nonoverlap split for ToTTo), and each sample is provided with all the inputs, the reference, and two system generated sentences. We report the percentage of faithful and adequate sentences generated by the baseline system and our system on all datasets in Table \ref{tab:human-eval}, and the results validate \texttt{R2D2}'s effectiveness in faithful text generation. 

\begin{table}[]
\centering
\resizebox{\hsize}{!}{
\begin{tabular}{@{}ccccc@{}}
\toprule
\textbf{Dataset} & \textbf{Systems} & \textbf{Split} & \textbf{Faithfulness (\%)} & \textbf{\begin{tabular}[c]{@{}c@{}}Coverage\\ w.r.t Ref (\%)\end{tabular}} \\
\midrule
\multirow{2}{*}{\textbf{FeTaQA}} & \textbf{Reg-FT} & A & 61.33 & 54.83 \\
 & \textbf{\texttt{R2D2}-FT} & A & \textbf{68.67} & \textbf{61.83} \\
 \noalign{\vskip 0.7ex}\hdashline\noalign{\vskip 0.7ex}
\multirow{2}{*}{\textbf{LogicNLG}} & \textbf{Reg-FT} & A & 40.67 & 69 \\
 & \textbf{\texttt{R2D2}-FT} & A & \textbf{41.17} & 66.50 \\
 \noalign{\vskip 0.7ex}\hdashline\noalign{\vskip 0.7ex}
\multirow{6}{*}{\textbf{ToTTo}} & \multirow{3}{*}{\textbf{Reg-FT}} & A & 81 & 81.17 \\
 &  & O & 83.66 & 82.66 \\
 &  & N & 78.32 & 79.66 \\
 & \multirow{3}{*}{\textbf{\texttt{R2D2-FT}}} & A & \textbf{83.16} & \textbf{84.67} \\
 &  & O & \textbf{84.34} & \textbf{88.66} \\
 &  & N & \textbf{82} & \textbf{80.66} \\
 \bottomrule
\end{tabular}
}%
\caption{Human evaluation result. A, O, N stands for All, Overlap and Nonoverlap.}
\label{tab:human-eval}
\end{table}

\subsection{Further Analysis}
\label{analysis}
Comparing the effectiveness of \texttt{R2D2} on different datasets and with different evaluations, we found that FeTaQA and ToTTo benefit more than LogicNLG. We speculate this is because many sentences of LogicNLG describe some entailed facts of entities of a single table column (usually involving comparisons), which usually contain single and less restricted predicate that could be applied to many homogeneous entities, and this would invalidate our perturbation methods. We provide one such example in Figure \ref{fig:invalidate-perturb} of the Appendix. We also notice that for ToTTo, the improvement is less evident than that for FeTaQA. Besides less room for improvement, we observe no evident change in the entities retrieved by both systems compared with those in the reference or input, while human evaluation indicates there are still around 17\% unfaithful sentences. We speculate the source of unfaithfulness of these sentences are due to wrong predictions of relations/predicates, which are not captured and included into the \texttt{R2D2} fine-tuning by our current perturbation method. To avoid invalidation of perturbation (as in the case of LogicNLG) and also to capture erroneous relation predictions, a better perturbation method has to operate on fact triples instead of entities, but this requires a reliable and domain-independent fact extraction system, which we will explore in future. 

\section{Related Work}


\subsection{Unfaithfulness in Text Generation}

In the context of text generation, hallucination refers to the phenomenon of neural models “generating unfaithful or nonsensical text” \citep{ji_et_al_2022_survey}. For instance, the generated text can contradict information in the source, or have nothing to do with the source, and hence cannot be verified. One reason for such hallucination is poor data collection; for example, information in the target sentence may not actually be present in the table, which can occur when the tables and sentences do not actually align. Another source of such hallucination is training, such as the exposure bias, or the task expects more output diversity. Metrics based on information extraction, question answering, and natural language inference, have been proposed to measure such hallucination, which we employ to evaluate the performance and faithfulness of \texttt{R2D2}.

\subsection{Contrastive Learning}

Contrastive learning~\citep{10.1109/CVPR.2006.100} tasks the model with maximizing the representation similarity between neighboring examples while minimizing the similarity between distant examples.
Contrastive learning has recently been used in various NLP tasks, including sentence embedding~\citep{gao-etal-2021-simcse}, machine translation~\citep{yang-etal-2019-reducing, pan-etal-2021-contrastive}, text summarization~\citep{cao-wang-2021-cliff, liu-liu-2021-simcls, DBLP:journals/corr/abs-2109-03481, DBLP:journals/corr/abs-2108-11846, wang-etal-2021-contrastive, Liu2022BRIOBO}, 
data-to-text generation~\citep{uehara-etal-2020-learning} and other tasks~\citep{cho-etal-2021-contrastive, lee2021contrastive, Su2022ACF, Su2021TaCLIB}. Unlike \citet{uehara-etal-2020-learning}, in which unfaithful sentences are obtained by replacing a set of keywords (such as replacing \textit{low} to \textit{high}, \textit{gain} to \textit{drop}) that only apply to the finance domain, we propose domain-independent methods for sampling unfaithful sentences either by exploiting the structure of input knowledge or utilizing the D2T model's own mistakes.

\subsection{Unlikelihood Training}
To address the degeneration problems of models trained only with Maximum Likelihood Estimation, many works have proposed alternative approaches \cite{tu-etal-2016-modeling, li-etal-2020-dont, holtzman_2020_degeneration, lin_etal_2021-straight}. Among them, unlikelihood training was introduced as a means of decreasing the probability that the model generates certain tokens \cite{welleck_et_al_2019_neural}. In a D2T context, we adopt unlikelihood training to decrease the probability that the model generates tokens which are not entailed by the given contexts. 

\subsection{Evaluation Metrics}

Ideally, a data-to-text model should be evaluated based on its ability to generate logical sentences based on the provided reference data. Current methods however, typically only compare the model output summary to the gold standard label. This includes n-gram based (e.g. BLEU, ROUGE, and METEOR) or edit distance based metrics (e.g. TER) \cite{sai-2022-survey-metrics}, or embedding-based similarity metrics (e.g. BERTScore) \cite{zhang-2019-bertscore}. Another set of metrics compare the information present in the output and the label. This is done by extracting subject, object, and their relations in the output and label, and comparing both sets of elements \cite{wiseman-etal-2017-challenges}. We evaluate our model using multiple metrics to understand different aspects of its performance. 

\subsection{Natural Language Inference}

Natural language inference (NLI) refers to the task of classifying whether a hypothesis entails, contradicts, or is unrelated to a premise \cite{bowman-2015-snli}. Benchmarks for this task include the Stanford NLI dataset (SNLI) \cite{bowman-2015-snli}, the Question-Answering NLI dataset (QNLI) \cite{wang-2018-glue}, and MultiNLI \cite{williams-2017-multinli}. In the context of D2T, NLI can be used to evaluate whether a model’s generated text can be inferred from the input table \cite{chen-etal-2020-logical}. An ideal model would generate sentences which an NLI model would classify as being entailed by the input table. In line with the work of TabFact \cite{chen_et_al_2019_tabfact}, LogicNLG \cite{chen-etal-2020-logical}, and SnowBall \cite{shu-etal-2021-logic}, \texttt{R2D2} incorporates this idea into data-to-text training by using NLI as a learning objective during the training procedure.

\section{Conclusion}
In this work, we introduced \texttt{R2D2}, a training framework that effectively mitigates the unfaithful text generation problem for the D2T task. Training with the regular maximum likelihood loss can lead to generation of sentences that are similar to the references but are unfaithful to the input. We therefore propose to add a discrimination task and an unlikelihood training to encourage the model to generate separable representations of these critical sentences. We proposed two methods of sampling these critical sentences that are unfaithful: the knowledge-based method exploits the structure of the input knowledge, and the model-based method samples the D2T model's own mistakes. We examined existing metrics for evaluating faithfulness of generated sentences, and proposed NER-based metrics that assess the entity retrieval capability of the Data-to-Text systems, as we argued the incompetence of which is one of the leading causes of unfaithfulness. We experimented on multiple Data-to-Text datasets of different task constructs, and achieved noticeable improvements over the state-of-the-art.


\bibliography{anthology,custom}

\begin{thebibliography}{56}
\expandafter\ifx\csname natexlab\endcsname\relax\def\natexlab#1{#1}\fi

\bibitem[{Banerjee and Lavie(2005)}]{banerjee-lavie-2005-meteor}
Satanjeev Banerjee and Alon Lavie. 2005.
\newblock \href {https://aclanthology.org/W05-0909} {{METEOR}: An automatic
  metric for {MT} evaluation with improved correlation with human judgments}.
\newblock In \emph{Proceedings of the {ACL} Workshop on Intrinsic and Extrinsic
  Evaluation Measures for Machine Translation and/or Summarization}, pages
  65--72, Ann Arbor, Michigan. Association for Computational Linguistics.

\bibitem[{Beltagy et~al.(2020)Beltagy, Peters, and
  Cohan}]{beltagy-etal-2020-longformer}
Iz~Beltagy, Matthew~E. Peters, and Arman Cohan. 2020.
\newblock Longformer: The long-document transformer.
\newblock \emph{arXiv:2004.05150}.

\bibitem[{Bowman et~al.(2015)Bowman, Angeli, Potts, and
  Manning}]{bowman-2015-snli}
Samuel~R. Bowman, Gabor Angeli, Christopher Potts, and Christopher~D. Manning.
  2015.
\newblock \href {https://doi.org/10.48550/ARXIV.1508.05326} {A large annotated
  corpus for learning natural language inference}.

\bibitem[{Cao and Wang(2021)}]{cao-wang-2021-cliff}
Shuyang Cao and Lu~Wang. 2021.
\newblock \href {https://doi.org/10.18653/v1/2021.emnlp-main.532} {{CLIFF}:
  Contrastive learning for improving faithfulness and factuality in abstractive
  summarization}.
\newblock In \emph{Proceedings of the 2021 Conference on Empirical Methods in
  Natural Language Processing}, pages 6633--6649, Online and Punta Cana,
  Dominican Republic. Association for Computational Linguistics.

\bibitem[{Cao et~al.(2017)Cao, Wei, Li, and Li}]{cao_et_al_2017_faithful}
Ziqiang Cao, Furu Wei, Wenjie Li, and Sujian Li. 2017.
\newblock \href {https://doi.org/10.48550/ARXIV.1711.04434} {Faithful to the
  original: Fact aware neural abstractive summarization}.

\bibitem[{Chen et~al.(2021{\natexlab{a}})Chen, Zhang, Sone, and
  Roth}]{chen-etal-2021-improving}
Sihao Chen, Fan Zhang, Kazoo Sone, and Dan Roth. 2021{\natexlab{a}}.
\newblock \href {https://doi.org/10.18653/v1/2021.naacl-main.475} {Improving
  faithfulness in abstractive summarization with contrast candidate generation
  and selection}.
\newblock In \emph{Proceedings of the 2021 Conference of the North American
  Chapter of the Association for Computational Linguistics: Human Language
  Technologies}, pages 5935--5941, Online. Association for Computational
  Linguistics.

\bibitem[{Chen et~al.(2020{\natexlab{a}})Chen, Chen, Su, Chen, and
  Wang}]{chen-etal-2020-logical}
Wenhu Chen, Jianshu Chen, Yu~Su, Zhiyu Chen, and William~Yang Wang.
  2020{\natexlab{a}}.
\newblock \href {https://doi.org/10.18653/v1/2020.acl-main.708} {Logical
  natural language generation from open-domain tables}.
\newblock In \emph{Proceedings of the 58th Annual Meeting of the Association
  for Computational Linguistics}, pages 7929--7942, Online. Association for
  Computational Linguistics.

\bibitem[{Chen et~al.(2020{\natexlab{b}})Chen, Wang, Chen, Zhang, Wang, Li,
  Zhou, and Wang}]{chen_et_al_2019_tabfact}
Wenhu Chen, Hongmin Wang, Jianshu Chen, Yunkai Zhang, Hong Wang, Shiyang Li,
  Xiyou Zhou, and William~Yang Wang. 2020{\natexlab{b}}.
\newblock \href {https://arxiv.org/abs/1909.02164} {Tabfact : A large-scale
  dataset for table-based fact verification}.
\newblock In \emph{International Conference on Learning Representations
  (ICLR)}, Addis Ababa, Ethiopia.

\bibitem[{Chen et~al.(2021{\natexlab{b}})Chen, Tian, Li, He, and
  Jin}]{chen-etal-2021-de}
Wenqing Chen, Jidong Tian, Yitian Li, Hao He, and Yaohui Jin.
  2021{\natexlab{b}}.
\newblock \href {https://doi.org/10.18653/v1/2021.acl-long.430} {De-confounded
  variational encoder-decoder for logical table-to-text generation}.
\newblock In \emph{Proceedings of the 59th Annual Meeting of the Association
  for Computational Linguistics and the 11th International Joint Conference on
  Natural Language Processing (Volume 1: Long Papers)}, pages 5532--5542,
  Online. Association for Computational Linguistics.

\bibitem[{Cho et~al.(2021)Cho, Zhang, Rao, Celikyilmaz, Xiong, Gao, Wang, and
  Dolan}]{cho-etal-2021-contrastive}
Woon~Sang Cho, Yizhe Zhang, Sudha Rao, Asli Celikyilmaz, Chenyan Xiong,
  Jianfeng Gao, Mengdi Wang, and Bill Dolan. 2021.
\newblock \href {https://doi.org/10.18653/v1/2021.eacl-main.2} {Contrastive
  multi-document question generation}.
\newblock In \emph{Proceedings of the 16th Conference of the European Chapter
  of the Association for Computational Linguistics: Main Volume}, pages 12--30,
  Online. Association for Computational Linguistics.

\bibitem[{Clark et~al.(2020)Clark, Luong, Le, and
  Manning}]{clark_et_al_2020_electra}
Kevin Clark, Minh-Thang Luong, Quoc~V. Le, and Christopher~D. Manning. 2020.
\newblock \href {https://openreview.net/pdf?id=r1xMH1BtvB} {{ELECTRA}:
  Pre-training text encoders as discriminators rather than generators}.
\newblock In \emph{ICLR}.

\bibitem[{Dhingra et~al.(2019)Dhingra, Faruqui, Parikh, Chang, Das, and
  Cohen}]{dhingra-etal-2019-handling}
Bhuwan Dhingra, Manaal Faruqui, Ankur Parikh, Ming-Wei Chang, Dipanjan Das, and
  William Cohen. 2019.
\newblock \href {https://doi.org/10.18653/v1/P19-1483} {Handling divergent
  reference texts when evaluating table-to-text generation}.
\newblock In \emph{Proceedings of the 57th Annual Meeting of the Association
  for Computational Linguistics}, pages 4884--4895, Florence, Italy.
  Association for Computational Linguistics.

\bibitem[{Gao et~al.(2021)Gao, Yao, and Chen}]{gao-etal-2021-simcse}
Tianyu Gao, Xingcheng Yao, and Danqi Chen. 2021.
\newblock \href {https://doi.org/10.18653/v1/2021.emnlp-main.552} {{S}im{CSE}:
  Simple contrastive learning of sentence embeddings}.
\newblock In \emph{Proceedings of the 2021 Conference on Empirical Methods in
  Natural Language Processing}, pages 6894--6910, Online and Punta Cana,
  Dominican Republic. Association for Computational Linguistics.

\bibitem[{Gardent et~al.(2017)Gardent, Shimorina, Narayan, and
  Perez-Beltrachini}]{gardent-etal-2017-webnlg}
Claire Gardent, Anastasia Shimorina, Shashi Narayan, and Laura
  Perez-Beltrachini. 2017.
\newblock \href {https://doi.org/10.18653/v1/W17-3518} {The {W}eb{NLG}
  challenge: Generating text from {RDF} data}.
\newblock In \emph{Proceedings of the 10th International Conference on Natural
  Language Generation}, pages 124--133, Santiago de Compostela, Spain.
  Association for Computational Linguistics.

\bibitem[{Gatt and Krahmer(2018)}]{gatt_krahmer_2018_survey}
Albert Gatt and Emiel Krahmer. 2018.
\newblock \href {https://arxiv.org/abs/1703.09902} {Survey of the state of the
  art in natural language generation: Core tasks, applications and evaluation}.
\newblock \emph{J. Artif. Int. Res.}, 61(1):65–170.

\bibitem[{Goyal and Durrett(2020)}]{goyal-durrett-2020-evaluating}
Tanya Goyal and Greg Durrett. 2020.
\newblock \href {https://doi.org/10.18653/v1/2020.findings-emnlp.322}
  {Evaluating factuality in generation with dependency-level entailment}.
\newblock In \emph{Findings of the Association for Computational Linguistics:
  EMNLP 2020}, pages 3592--3603, Online. Association for Computational
  Linguistics.

\bibitem[{Goyal and Durrett(2021)}]{goyal-durrett-2021-annotating}
Tanya Goyal and Greg Durrett. 2021.
\newblock \href {https://doi.org/10.18653/v1/2021.naacl-main.114} {Annotating
  and modeling fine-grained factuality in summarization}.
\newblock In \emph{Proceedings of the 2021 Conference of the North American
  Chapter of the Association for Computational Linguistics: Human Language
  Technologies}, pages 1449--1462, Online. Association for Computational
  Linguistics.

\bibitem[{Hadsell et~al.(2006)Hadsell, Chopra, and
  LeCun}]{10.1109/CVPR.2006.100}
Raia Hadsell, Sumit Chopra, and Yann LeCun. 2006.
\newblock \href {https://doi.org/10.1109/CVPR.2006.100} {Dimensionality
  reduction by learning an invariant mapping}.
\newblock In \emph{Proceedings of the 2006 IEEE Computer Society Conference on
  Computer Vision and Pattern Recognition - Volume 2}, CVPR '06, page
  1735–1742, USA. IEEE Computer Society.

\bibitem[{Holtzman et~al.(2020)Holtzman, Buys, Du, Forbes, and
  Choi}]{holtzman_2020_degeneration}
Ari Holtzman, Jan Buys, Li~Du, Maxwell Forbes, and Yejin Choi. 2020.
\newblock \href {https://openreview.net/forum?id=rygGQyrFvH} {The curious case
  of neural text degeneration}.
\newblock In \emph{International Conference on Learning Representations}.

\bibitem[{Ji et~al.(2022)Ji, Lee, Frieske, Yu, Su, Xu, Ishii, Bang, Madotto,
  and Fung}]{ji_et_al_2022_survey}
Ziwei Ji, Nayeon Lee, Rita Frieske, Tiezheng Yu, Dan Su, Yan Xu, Etsuko Ishii,
  Yejin Bang, Andrea Madotto, and Pascale Fung. 2022.
\newblock \href {https://doi.org/10.48550/ARXIV.2202.03629} {Survey of
  hallucination in natural language generation}.

\bibitem[{Kale and Rastogi(2020)}]{kale-rastogi-2020-text}
Mihir Kale and Abhinav Rastogi. 2020.
\newblock \href {https://aclanthology.org/2020.inlg-1.14} {Text-to-text
  pre-training for data-to-text tasks}.
\newblock In \emph{Proceedings of the 13th International Conference on Natural
  Language Generation}, pages 97--102, Dublin, Ireland. Association for
  Computational Linguistics.

\bibitem[{Lebret et~al.(2016)Lebret, Grangier, and
  Auli}]{lebret-etal-2016-neural}
R{\'e}mi Lebret, David Grangier, and Michael Auli. 2016.
\newblock \href {https://doi.org/10.18653/v1/D16-1128} {Neural text generation
  from structured data with application to the biography domain}.
\newblock In \emph{Proceedings of the 2016 Conference on Empirical Methods in
  Natural Language Processing}, pages 1203--1213, Austin, Texas. Association
  for Computational Linguistics.

\bibitem[{Lee et~al.(2021)Lee, Lee, and Hwang}]{lee2021contrastive}
Seanie Lee, Dong~Bok Lee, and Sung~Ju Hwang. 2021.
\newblock \href {https://openreview.net/forum?id=Wga_hrCa3P3} {Contrastive
  learning with adversarial perturbations for conditional text generation}.
\newblock In \emph{International Conference on Learning Representations}.

\bibitem[{Lewis et~al.(2020)Lewis, Liu, Goyal, Ghazvininejad, Mohamed, Levy,
  Stoyanov, and Zettlemoyer}]{lewis-etal-2020-bart}
Mike Lewis, Yinhan Liu, Naman Goyal, Marjan Ghazvininejad, Abdelrahman Mohamed,
  Omer Levy, Veselin Stoyanov, and Luke Zettlemoyer. 2020.
\newblock \href {https://doi.org/10.18653/v1/2020.acl-main.703} {{BART}:
  Denoising sequence-to-sequence pre-training for natural language generation,
  translation, and comprehension}.
\newblock In \emph{Proceedings of the 58th Annual Meeting of the Association
  for Computational Linguistics}, pages 7871--7880, Online. Association for
  Computational Linguistics.

\bibitem[{Li et~al.(2020)Li, Roller, Kulikov, Welleck, Boureau, Cho, and
  Weston}]{li-etal-2020-dont}
Margaret Li, Stephen Roller, Ilia Kulikov, Sean Welleck, Y-Lan Boureau,
  Kyunghyun Cho, and Jason Weston. 2020.
\newblock \href {https://doi.org/10.18653/v1/2020.acl-main.428} {Don{'}t say
  that! making inconsistent dialogue unlikely with unlikelihood training}.
\newblock In \emph{Proceedings of the 58th Annual Meeting of the Association
  for Computational Linguistics}, pages 4715--4728, Online. Association for
  Computational Linguistics.

\bibitem[{Lin(2004)}]{lin-2004-rouge}
Chin-Yew Lin. 2004.
\newblock \href {https://aclanthology.org/W04-1013} {{ROUGE}: A package for
  automatic evaluation of summaries}.
\newblock In \emph{Text Summarization Branches Out}, pages 74--81, Barcelona,
  Spain. Association for Computational Linguistics.

\bibitem[{Lin et~al.(2021)Lin, Han, and Joty}]{lin_etal_2021-straight}
Xiang Lin, Simeng Han, and Shafiq Joty. 2021.
\newblock \href {https://proceedings.mlr.press/v139/lin21b.html} {Straight to
  the gradient: Learning to use novel tokens for neural text generation}.
\newblock In \emph{Proceedings of the 38th International Conference on Machine
  Learning}, volume 139 of \emph{Proceedings of Machine Learning Research},
  pages 6642--6653. PMLR.

\bibitem[{Liu and Liu(2021)}]{liu-liu-2021-simcls}
Yixin Liu and Pengfei Liu. 2021.
\newblock \href {https://doi.org/10.18653/v1/2021.acl-short.135} {{S}im{CLS}: A
  simple framework for contrastive learning of abstractive summarization}.
\newblock In \emph{Proceedings of the 59th Annual Meeting of the Association
  for Computational Linguistics and the 11th International Joint Conference on
  Natural Language Processing (Volume 2: Short Papers)}, pages 1065--1072,
  Online. Association for Computational Linguistics.

\bibitem[{Liu et~al.(2022)Liu, Liu, Radev, and Neubig}]{Liu2022BRIOBO}
Yixin Liu, Pengfei Liu, Dragomir Radev, and Graham Neubig. 2022.
\newblock \href {https://aclanthology.org/2022.acl-long.207} {{BRIO}: Bringing
  order to abstractive summarization}.
\newblock In \emph{Proceedings of the 60th Annual Meeting of the Association
  for Computational Linguistics (Volume 1: Long Papers)}, pages 2890--2903,
  Dublin, Ireland. Association for Computational Linguistics.

\bibitem[{Maynez et~al.(2020)Maynez, Narayan, Bohnet, and
  McDonald}]{maynez-etal-2020-faithfulness}
Joshua Maynez, Shashi Narayan, Bernd Bohnet, and Ryan McDonald. 2020.
\newblock \href {https://doi.org/10.18653/v1/2020.acl-main.173} {On
  faithfulness and factuality in abstractive summarization}.
\newblock In \emph{Proceedings of the 58th Annual Meeting of the Association
  for Computational Linguistics}, pages 1906--1919, Online. Association for
  Computational Linguistics.

\bibitem[{Nan et~al.(2022)Nan, Hsieh, Mao, Lin, Verma, Zhang, Kryściński,
  Schoelkopf, Kong, Tang, Mutuma, Rosand, Trindade, Bandaru, Cunningham, Xiong,
  and Radev}]{nan_et_al_2022_fetaqa}
Linyong Nan, Chiachun Hsieh, Ziming Mao, Xi~Victoria Lin, Neha Verma, Rui
  Zhang, Wojciech Kryściński, Nick Schoelkopf, Riley Kong, Xiangru Tang,
  Mutethia Mutuma, Ben Rosand, Isabel Trindade, Renusree Bandaru, Jacob
  Cunningham, Caiming Xiong, and Dragomir Radev. 2022.
\newblock \href {https://doi.org/10.1162/tacl_a_00446} {{FeTaQA: Free-form
  Table Question Answering}}.
\newblock \emph{Transactions of the Association for Computational Linguistics},
  10:35--49.

\bibitem[{Novikova et~al.(2017)Novikova, Du{\v{s}}ek, and
  Rieser}]{novikova-etal-2017-e2e}
Jekaterina Novikova, Ond{\v{r}}ej Du{\v{s}}ek, and Verena Rieser. 2017.
\newblock \href {https://doi.org/10.18653/v1/W17-5525} {The {E}2{E} dataset:
  New challenges for end-to-end generation}.
\newblock In \emph{Proceedings of the 18th Annual {SIG}dial Meeting on
  Discourse and Dialogue}, pages 201--206, Saarbr{\"u}cken, Germany.
  Association for Computational Linguistics.

\bibitem[{Pan et~al.(2021)Pan, Wang, Wu, and Li}]{pan-etal-2021-contrastive}
Xiao Pan, Mingxuan Wang, Liwei Wu, and Lei Li. 2021.
\newblock \href {https://doi.org/10.18653/v1/2021.acl-long.21} {Contrastive
  learning for many-to-many multilingual neural machine translation}.
\newblock In \emph{Proceedings of the 59th Annual Meeting of the Association
  for Computational Linguistics and the 11th International Joint Conference on
  Natural Language Processing (Volume 1: Long Papers)}, pages 244--258, Online.
  Association for Computational Linguistics.

\bibitem[{Parikh et~al.(2020)Parikh, Wang, Gehrmann, Faruqui, Dhingra, Yang,
  and Das}]{parikh-etal-2020-totto}
Ankur Parikh, Xuezhi Wang, Sebastian Gehrmann, Manaal Faruqui, Bhuwan Dhingra,
  Diyi Yang, and Dipanjan Das. 2020.
\newblock \href {https://doi.org/10.18653/v1/2020.emnlp-main.89} {{ToTTo}: A
  controlled table-to-text generation dataset}.
\newblock In \emph{Proceedings of the 2020 Conference on Empirical Methods in
  Natural Language Processing (EMNLP)}, pages 1173--1186, Online. Association
  for Computational Linguistics.

\bibitem[{Post(2018)}]{post-2018-call}
Matt Post. 2018.
\newblock \href {https://doi.org/10.18653/v1/W18-6319} {A call for clarity in
  reporting {BLEU} scores}.
\newblock In \emph{Proceedings of the Third Conference on Machine Translation:
  Research Papers}, pages 186--191, Brussels, Belgium. Association for
  Computational Linguistics.

\bibitem[{Raffel et~al.(2020)Raffel, Shazeer, Roberts, Lee, Narang, Matena,
  Zhou, Li, and Liu}]{raffel-etal-2020-t5}
Colin Raffel, Noam Shazeer, Adam Roberts, Katherine Lee, Sharan Narang, Michael
  Matena, Yanqi Zhou, Wei Li, and Peter~J. Liu. 2020.
\newblock \href {http://jmlr.org/papers/v21/20-074.html} {Exploring the limits
  of transfer learning with a unified text-to-text transformer}.
\newblock \emph{Journal of Machine Learning Research}, 21(140):1--67.

\bibitem[{Reiter and Dale(2000)}]{reiter_dale_2000_studies}
Ehud Reiter and Robert Dale. 2000.
\newblock \href {https://doi.org/10.1017/CBO9780511519857} {\emph{Building
  Natural Language Generation Systems}}.
\newblock Studies in Natural Language Processing. Cambridge University Press.

\bibitem[{Sai et~al.(2022)Sai, Mohankumar, and
  Khapra}]{sai-2022-survey-metrics}
Ananya~B. Sai, Akash~Kumar Mohankumar, and Mitesh~M. Khapra. 2022.
\newblock \href {https://doi.org/10.1145/3485766} {A survey of evaluation
  metrics used for nlg systems}.
\newblock \emph{ACM Comput. Surv.}, 55(2).

\bibitem[{Shazeer and Stern(2018)}]{shazeer-stern-2018-adafactor}
Noam Shazeer and Mitchell Stern. 2018.
\newblock \href {https://proceedings.mlr.press/v80/shazeer18a.html} {Adafactor:
  Adaptive learning rates with sublinear memory cost}.
\newblock In \emph{Proceedings of the 35th International Conference on Machine
  Learning}, volume~80 of \emph{Proceedings of Machine Learning Research},
  pages 4596--4604. PMLR.

\bibitem[{Shu et~al.(2021)Shu, Zhang, Dong, Shi, Yu, and
  Zhang}]{shu-etal-2021-logic}
Chang Shu, Yusen Zhang, Xiangyu Dong, Peng Shi, Tao Yu, and Rui Zhang. 2021.
\newblock \href {https://doi.org/10.18653/v1/2021.findings-acl.388}
  {Logic-consistency text generation from semantic parses}.
\newblock In \emph{Findings of the Association for Computational Linguistics:
  ACL-IJCNLP 2021}, pages 4414--4426, Online. Association for Computational
  Linguistics.

\bibitem[{Snover et~al.(2006)Snover, Dorr, Schwartz, Micciulla, and
  Makhoul}]{snover-etal-2006-study}
Matthew Snover, Bonnie Dorr, Rich Schwartz, Linnea Micciulla, and John Makhoul.
  2006.
\newblock \href {https://aclanthology.org/2006.amta-papers.25} {A study of
  translation edit rate with targeted human annotation}.
\newblock In \emph{Proceedings of the 7th Conference of the Association for
  Machine Translation in the Americas: Technical Papers}, pages 223--231,
  Cambridge, Massachusetts, USA. Association for Machine Translation in the
  Americas.

\bibitem[{Su et~al.(2022)Su, Lan, Wang, Yogatama, Kong, and
  Collier}]{Su2022ACF}
Yixuan Su, Tian Lan, Yan Wang, Dani Yogatama, Lingpeng Kong, and Nigel Collier.
  2022.
\newblock \href {https://arxiv.org/abs/2202.06417} {A contrastive framework for
  neural text generation}.
\newblock \emph{ArXiv}, abs/2202.06417.

\bibitem[{Su et~al.(2021)Su, Liu, Meng, Shu, Shareghi, and
  Collier}]{Su2021TaCLIB}
Yixuan Su, Fangyu Liu, Zaiqiao Meng, Lei Shu, Ehsan Shareghi, and Nigel
  Collier. 2021.
\newblock \href {https://arxiv.org/abs/2111.04198} {Tacl: Improving bert
  pre-training with token-aware contrastive learning}.
\newblock \emph{ArXiv}, abs/2111.04198.

\bibitem[{Sun and Li(2021)}]{DBLP:journals/corr/abs-2108-11846}
Shichao Sun and Wenjie Li. 2021.
\newblock \href {http://arxiv.org/abs/2108.11846} {Alleviating exposure bias
  via contrastive learning for abstractive text summarization}.
\newblock \emph{CoRR}, abs/2108.11846.

\bibitem[{Tu et~al.(2016)Tu, Lu, Liu, Liu, and Li}]{tu-etal-2016-modeling}
Zhaopeng Tu, Zhengdong Lu, Yang Liu, Xiaohua Liu, and Hang Li. 2016.
\newblock \href {https://doi.org/10.18653/v1/P16-1008} {Modeling coverage for
  neural machine translation}.
\newblock In \emph{Proceedings of the 54th Annual Meeting of the Association
  for Computational Linguistics (Volume 1: Long Papers)}, pages 76--85, Berlin,
  Germany. Association for Computational Linguistics.

\bibitem[{Uehara et~al.(2020)Uehara, Ishigaki, Aoki, Noji, Goshima, Kobayashi,
  Takamura, and Miyao}]{uehara-etal-2020-learning}
Yui Uehara, Tatsuya Ishigaki, Kasumi Aoki, Hiroshi Noji, Keiichi Goshima,
  Ichiro Kobayashi, Hiroya Takamura, and Yusuke Miyao. 2020.
\newblock \href {https://doi.org/10.18653/v1/2020.coling-main.213} {Learning
  with contrastive examples for data-to-text generation}.
\newblock In \emph{Proceedings of the 28th International Conference on
  Computational Linguistics}, pages 2352--2362, Barcelona, Spain (Online).
  International Committee on Computational Linguistics.

\bibitem[{Wang et~al.(2018)Wang, Singh, Michael, Hill, Levy, and
  Bowman}]{wang-2018-glue}
Alex Wang, Amanpreet Singh, Julian Michael, Felix Hill, Omer Levy, and Samuel
  Bowman. 2018.
\newblock \href {https://doi.org/10.18653/v1/W18-5446} {{GLUE}: A multi-task
  benchmark and analysis platform for natural language understanding}.
\newblock In \emph{Proceedings of the 2018 {EMNLP} Workshop {B}lackbox{NLP}:
  Analyzing and Interpreting Neural Networks for {NLP}}, pages 353--355,
  Brussels, Belgium. Association for Computational Linguistics.

\bibitem[{Wang et~al.(2021)Wang, Chen, Zhou, Qiu, and
  Li}]{wang-etal-2021-contrastive}
Danqing Wang, Jiaze Chen, Hao Zhou, Xipeng Qiu, and Lei Li. 2021.
\newblock \href {https://doi.org/10.18653/v1/2021.findings-acl.242}
  {Contrastive aligned joint learning for multilingual summarization}.
\newblock In \emph{Findings of the Association for Computational Linguistics:
  ACL-IJCNLP 2021}, pages 2739--2750, Online. Association for Computational
  Linguistics.

\bibitem[{Welleck et~al.(2019)Welleck, Kulikov, Roller, Dinan, Cho, and
  Weston}]{welleck_et_al_2019_neural}
Sean Welleck, Ilia Kulikov, Stephen Roller, Emily Dinan, Kyunghyun Cho, and
  Jason Weston. 2019.
\newblock \href {http://arxiv.org/abs/1908.04319} {Neural text generation with
  unlikelihood training}.

\bibitem[{Williams et~al.(2017)Williams, Nangia, and
  Bowman}]{williams-2017-multinli}
Adina Williams, Nikita Nangia, and Samuel~R. Bowman. 2017.
\newblock \href {https://doi.org/10.48550/ARXIV.1704.05426} {A broad-coverage
  challenge corpus for sentence understanding through inference}.

\bibitem[{Wiseman et~al.(2017)Wiseman, Shieber, and
  Rush}]{wiseman-etal-2017-challenges}
Sam Wiseman, Stuart Shieber, and Alexander Rush. 2017.
\newblock \href {https://doi.org/10.18653/v1/D17-1239} {Challenges in
  data-to-document generation}.
\newblock In \emph{Proceedings of the 2017 Conference on Empirical Methods in
  Natural Language Processing}, pages 2253--2263, Copenhagen, Denmark.
  Association for Computational Linguistics.

\bibitem[{Xie et~al.(2022)Xie, Wu, Shi, Zhong, Scholak, Yasunaga, Wu, Zhong,
  Yin, Wang, Zhong, Wang, Li, Boyle, Ni, Yao, Radev, Xiong, Kong, Zhang, Smith,
  Zettlemoyer, and Yu}]{xie_et_al_2022_unifiedskg}
Tianbao Xie, Chen~Henry Wu, Peng Shi, Ruiqi Zhong, Torsten Scholak, Michihiro
  Yasunaga, Chien-Sheng Wu, Ming Zhong, Pengcheng Yin, Sida~I. Wang, Victor
  Zhong, Bailin Wang, Chengzu Li, Connor Boyle, Ansong Ni, Ziyu Yao, Dragomir
  Radev, Caiming Xiong, Lingpeng Kong, Rui Zhang, Noah~A. Smith, Luke
  Zettlemoyer, and Tao Yu. 2022.
\newblock \href {https://arxiv.org/abs/2201.05966} {Unifiedskg: Unifying and
  multi-tasking structured knowledge grounding with text-to-text language
  models}.
\newblock \emph{arXiv preprint arXiv:2201.05966}.

\bibitem[{Xu et~al.(2021)Xu, Zhang, Wu, and
  Wei}]{DBLP:journals/corr/abs-2109-03481}
Shusheng Xu, Xingxing Zhang, Yi~Wu, and Furu Wei. 2021.
\newblock \href {http://arxiv.org/abs/2109.03481} {Sequence level contrastive
  learning for text summarization}.
\newblock \emph{CoRR}, abs/2109.03481.

\bibitem[{Yang et~al.(2019)Yang, Cheng, Liu, and Sun}]{yang-etal-2019-reducing}
Zonghan Yang, Yong Cheng, Yang Liu, and Maosong Sun. 2019.
\newblock \href {https://doi.org/10.18653/v1/P19-1623} {Reducing word omission
  errors in neural machine translation: A contrastive learning approach}.
\newblock In \emph{Proceedings of the 57th Annual Meeting of the Association
  for Computational Linguistics}, pages 6191--6196, Florence, Italy.
  Association for Computational Linguistics.

\bibitem[{Zhang et~al.(2019)Zhang, Kishore, Wu, Weinberger, and
  Artzi}]{zhang-2019-bertscore}
Tianyi Zhang, Varsha Kishore, Felix Wu, Kilian~Q. Weinberger, and Yoav Artzi.
  2019.
\newblock \href {https://doi.org/10.48550/ARXIV.1904.09675} {Bertscore:
  Evaluating text generation with bert}.

\bibitem[{Zhang et~al.(2020)Zhang, Kishore, Wu, Weinberger, and
  Artzi}]{zhang-etal-2020-bertscore}
Tianyi Zhang, Varsha Kishore, Felix Wu, Kilian~Q. Weinberger, and Yoav Artzi.
  2020.
\newblock \href {https://openreview.net/forum?id=SkeHuCVFDr} {Bertscore:
  Evaluating text generation with bert}.
\newblock In \emph{International Conference on Learning Representations}.

\end{thebibliography}

\appendix

\section{Appendix}
\label{sec:appendix}

\begin{table*}[]
\centering
\resizebox{\hsize}{!}{
\begin{tabular}{@{\extracolsep{5pt}}ccccccccccc@{}}
\toprule
\multirow{2.5}{*}{\textbf{\begin{tabular}[c]{@{}c@{}}Discrimination\\ Loss\end{tabular}}} & \multirow{2.5}{*}{\textbf{\begin{tabular}[c]{@{}c@{}}Discrimination\\ Granularity\end{tabular}}} & \multirow{2.5}{*}{\textbf{\begin{tabular}[c]{@{}c@{}}Unlikelihood\\ Loss\end{tabular}}} & \multirow{2.5}{*}{\textbf{\begin{tabular}[c]{@{}c@{}}Perturbation\\ Method\end{tabular}}} & \multirow{2.5}{*}{\textbf{\begin{tabular}[c]{@{}c@{}}Perturbation\\ Size\end{tabular}}} & \textbf{Fact-based} & \multicolumn{5}{c}{\textbf{String-based}} \\
\cmidrule{6-6} \cmidrule{7-11} 
 &  &  &  &  & \textbf{NLI-Acc} & \textbf{sacreBLEU} & \textbf{Rouge-1/2/L} & \textbf{PARENT} & \textbf{TER} & \textbf{METEOR} \\
 \midrule
\multirow{5}{*}{\textbf{\cmark}} & \multirow{5}{*}{\textbf{sent-level}} & \multirow{5}{*}{\textbf{\xmark}} & \multirow{5}{*}{\textbf{\begin{tabular}[c]{@{}c@{}}Knowledge\\ based\end{tabular}}} & \texttt{xsmall} & 75.79 & 29.7 & 63.26/41.08/53.54 & 43.21 & 66.65 & 54.69 \\
 &  &  &  & \texttt{small} & 74.59 & 29.7 & 63.53/41.54/54.06 & 43.43 & 65.75 & 54.67 \\
 &  &  &  & \texttt{medium} & 77.13 & 30.2 & 63.47/41.43/53.68 & 43.99 & 66.96 & 55.09 \\
 &  &  &  & \texttt{large} & 74.64 & 29.2 & 62.57/40.74/53.22 & 42.74 & 66.93 & 53.76 \\
 &  &  &  & \texttt{full} & 73.99 & 29.4 & 63.26/41.24153.83 & 43.45 & 66.58 & 54.32 \\
 \noalign{\vskip 0.7ex}\hdashline\noalign{\vskip 0.7ex}
\multirow{2}{*}{\textbf{\cmark}} & \multirow{2}{*}{\textbf{sent-level}} & \multirow{2}{*}{\textbf{\xmark}} & \multirow{2}{*}{\textbf{\begin{tabular}[c]{@{}c@{}}Model\\ based\end{tabular}}} & \texttt{xsmall} & 76.83 & 30.5 & 63.46/41.57/53.84 & 44.49 & 66.47 & 55.33 \\
 &  &  &  & \texttt{full} & 74.69 & 29.2 & 63.08/41.03/53.57 & 43.01 & 66.76 & 54.39 \\
 \noalign{\vskip 0.7ex}\hdashline\noalign{\vskip 0.7ex}
\multirow{5}{*}{\textbf{\cmark}} & \multirow{5}{*}{\textbf{sent-level}} & \multirow{5}{*}{\textbf{\cmark}} & \multirow{5}{*}{\textbf{\begin{tabular}[c]{@{}c@{}}Knowledge\\ based\end{tabular}}} & \texttt{xsmall} & 76.14 & 30.4 & 63.36/41.03/53.64 & 43.61 & 67.89 & 55.61 \\
 &  &  &  & \texttt{small} & 76.44 & 31.4 & 63.56/41.33/53.93 & 44.68 & 68.95 & 56.34 \\
 &  &  &  & \texttt{medium} & 76.59 & 30.7 & 63.11/40.75/53.58 & 43.83 & 69.73 & 55.86 \\
 &  &  &  & \texttt{large} & 78.33 & 31.2 & 63.41/41.21/53.87 & 44.87 & 69.37 & 56.57 \\
 &  &  &  & \texttt{full} & 78.18 & 30.9 & 62.95/40.95/53.26 & 44.45 & 71.78 & 55.81 \\
 \noalign{\vskip 0.7ex}\hdashline\noalign{\vskip 0.7ex}
\multirow{2}{*}{\textbf{\cmark}} & \multirow{2}{*}{\textbf{sent-level}} & \multirow{2}{*}{\textbf{\cmark}} & \multirow{2}{*}{\textbf{\begin{tabular}[c]{@{}c@{}}Model\\ based\end{tabular}}} & \texttt{xsmall} & 77.58 & 30.7 & 63.30/41.14/53.34 & 44.18 & 68.48 & 55.47 \\
 &  &  &  & \texttt{full} & 75.39 & 29.6 & 62.43/40.33/52.95 & 42.45 & 69.75 & 54.26 \\
 \noalign{\vskip 0.7ex}\hdashline\noalign{\vskip 0.7ex}
\multirow{5}{*}{\textbf{\cmark}} & \multirow{5}{*}{\textbf{token-level}} & \multirow{5}{*}{\textbf{\xmark}} & \multirow{5}{*}{\textbf{\begin{tabular}[c]{@{}c@{}}Knowledge\\ based\end{tabular}}} & \texttt{xsmall} & 74.64 & 30.0 & 63.01/40.38/53.16 & 43.15 & 68.68 & 55.46 \\
 &  &  &  & \texttt{small} & 75.74 & 30.8 & 63.98/41.92/54.39 & 44.35 & 66.81 & 55.75 \\
 &  &  &  & \texttt{medium} & 75.54 & 30.5 & 63.41/41.42/53.93 & 44.36 & 67.47 & 55.60 \\
 &  &  &  & \texttt{large} & 74.14 & 29.9 & 63.41/41.16/53.68 & 43.66 & 67.61 & 55.31 \\
 &  &  &  & \texttt{full} & 74.94 & 29.2 & 63.23/40.73/53.64 & 42.62 & 67.89 & 54.33 \\
 \noalign{\vskip 0.7ex}\hdashline\noalign{\vskip 0.7ex}
\multirow{2}{*}{\textbf{\cmark}} & \multirow{2}{*}{\textbf{token-level}} & \multirow{2}{*}{\textbf{\xmark}} & \multirow{2}{*}{\textbf{\begin{tabular}[c]{@{}c@{}}Model\\ based\end{tabular}}} & \texttt{xsmall} & 75.39 & 30.3 & 63.27/40.96/53.66 & 43.39 & 67.40 & 55.16 \\
 &  &  &  & \texttt{full} & 74.44 & 30.4 & 63.34/41.24/53.92 & 44.11 & 67.30 & 55.58 \\
 \noalign{\vskip 0.7ex}\hdashline\noalign{\vskip 0.7ex}
\multirow{5}{*}{\textbf{\cmark}} & \multirow{5}{*}{\textbf{token-level}} & \multirow{5}{*}{\textbf{\cmark}} & \multirow{5}{*}{\textbf{\begin{tabular}[c]{@{}c@{}}Knowledge\\ based\end{tabular}}} & \texttt{xsmall} & 75.79 & 30.7 & 63.52/41.02/53.75 & 43.83 & 68.09 & 56.19 \\
 &  &  &  & \texttt{small} & 77.73 & 31.2 & 63.30/41.01/53.54 & 44.55 & 69.52 & 56.26 \\
 &  &  &  & \texttt{medium} & 77.93 & 31.5 & 63.50/41.71/54.05 & 45.32 & 68.55 & 56.27 \\
 &  &  &  & \texttt{large} & 76.78 & 30.9 & 62.67/40.76/53.41 & 43.93 & 71.47 & 55.59 \\
 &  &  &  & \texttt{full} & 78.28 & 30.9 & 62.50/40.43/53.07 & 44.29 & 71.24 & 55.69 \\
 \noalign{\vskip 0.7ex}\hdashline\noalign{\vskip 0.7ex}
\multirow{2}{*}{\textbf{\cmark}} & \multirow{2}{*}{\textbf{token-level}} & \multirow{2}{*}{\textbf{\cmark}} & \multirow{2}{*}{\textbf{\begin{tabular}[c]{@{}c@{}}Model\\ based\end{tabular}}} & \texttt{xsmall} & 75.28 & 29.8 & 62.82/40.33/53.10 & 42.72 & 68.92 & 54.48 \\
 &  &  &  & \texttt{full} & 75.19 & 29.2 & 61.88/39.49/52.11 & 41.68 & 72.00 & 54.05 \\
 \noalign{\vskip 0.7ex}\hdashline\noalign{\vskip 0.7ex}
\multirow{5}{*}{\textbf{\xmark}} & \multirow{5}{*}{\textbf{-}} & \multirow{5}{*}{\textbf{\cmark}} & \multirow{5}{*}{\textbf{\begin{tabular}[c]{@{}c@{}}Knowledge\\ based\end{tabular}}} & \texttt{xsmall} & 75.39 & 30.0 & 63.00/40.41/53.49 & 42.85 & 67.95 & 54.93 \\
 &  &  &  & \texttt{small} & 77.93 & 31.1 & 63.74/41.68/54.26 & 44.59 & 68.43 & 56.11 \\
 &  &  &  & \texttt{medium} & 77.58 & 30.8 & 63.34/41.08/53.70 & 43.87 & 68.56 & 55.77 \\
 &  &  &  & \texttt{large} & 76.29 & 31.5 & 63.11/41.20/53.82 & 45.11 & 69.93 & 56.30 \\
 &  &  &  & \texttt{full} & 77.68 & 31.0 & 63.17/40.96/53.53 & 44.20 & 70.97 & 56.24 \\
 \noalign{\vskip 0.7ex}\hdashline\noalign{\vskip 0.7ex}
\multirow{2}{*}{\textbf{\xmark}} & \multirow{2}{*}{\textbf{-}} & \multirow{2}{*}{\textbf{\cmark}} & \multirow{2}{*}{\textbf{\begin{tabular}[c]{@{}c@{}}Model\\ based\end{tabular}}} & \texttt{xsmall} & 74.34 & 30.2 & 63.06/40.56/53.41 & 43.62 & 68.65 & 55.22 \\
 &  &  &  & \texttt{full} & 77.33 & 29.9 & 62.27/40.16/52.50 & 43.00 & 73.97 & 54.95 \\
 \bottomrule
\end{tabular}
}%
\caption{Full \texttt{R2D2} experiment results for FeTaQA.}
\label{tab:full-result}
\end{table*}

\begin{figure*}
  \centering
  \includegraphics[width=0.5\textwidth]{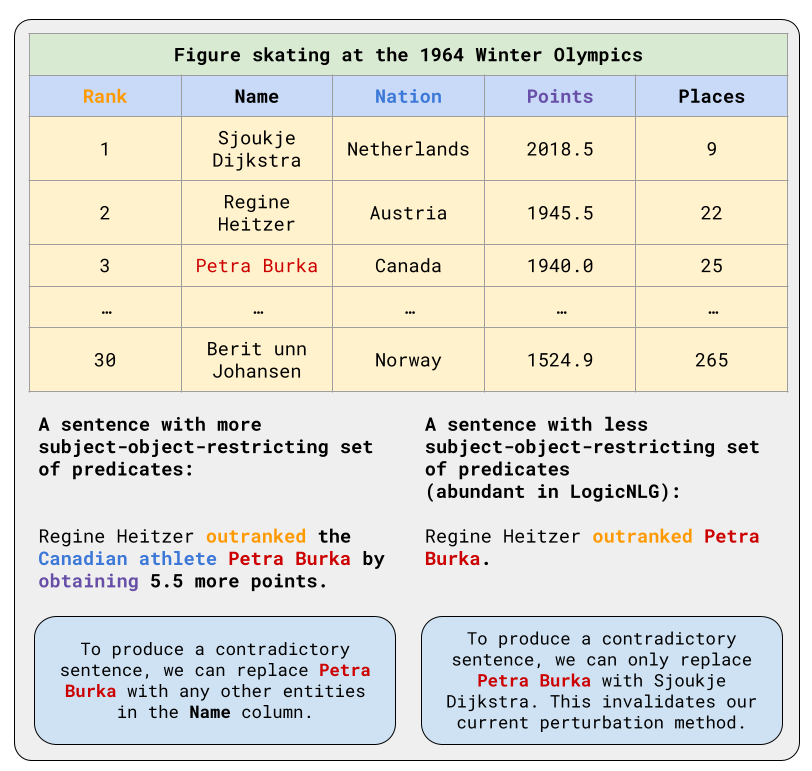}
  \caption{Sentences for which our current perturbation methods do not apply.}
  \label{fig:invalidate-perturb}
\end{figure*}

\end{document}